\documentclass[10pt,twocolumn,letterpaper]{article}

 \usepackage{cvpr}              %

\usepackage{graphicx}
\usepackage{amsmath}
\usepackage{amssymb}
\usepackage{booktabs}
\usepackage{mathtools}
\usepackage[accsupp]{axessibility}

\usepackage[pagebackref,breaklinks,colorlinks]{hyperref}

\usepackage[resetlabels,labeled]{multibib}
\newcites{A}{Additional References}

\usepackage[capitalize]{cleveref}
\crefname{section}{Sec.}{Secs.}
\Crefname{section}{Section}{Sections}
\Crefname{table}{Table}{Tables}
\crefname{table}{Tab.}{Tabs.}

\usepackage[dvipsnames]{xcolor}

\def\cvprPaperID{9831} %
\def\confName{CVPR}
\def\confYear{2022}

\usepackage{overpic}
\usepackage{enumitem} %
\usepackage{overpic} %
\usepackage{color}

\definecolor{turquoise}{cmyk}{0.65,0,0.1,0.3}
\definecolor{purple}{rgb}{0.65,0,0.65}
\definecolor{dark_green}{rgb}{0, 0.5, 0}
\definecolor{orange}{rgb}{0.8, 0.6, 0.2}
\definecolor{red}{rgb}{0.8, 0.2, 0.2}
\definecolor{darkred}{rgb}{0.6, 0.1, 0.05}
\definecolor{blueish}{rgb}{0.0, 0.3, .6}
\definecolor{light_gray}{rgb}{0.7, 0.7, .7}
\definecolor{pink}{rgb}{1, 0, 1}
\definecolor{greyblue}{rgb}{0.25, 0.25, 1}

\DeclareMathOperator*{\argmax}{arg\,max}

\usepackage{blindtext}

\renewcommand{\paragraph}[1]{\vspace{1em}\noindent\textbf{#1}.}

\begin{document}
\title{Joint Forecasting of Panoptic Segmentations with Difference Attention}

\author{Colin Graber$^1$~~~Cyril Jazra$^1$~~~Wenjie Luo$^2$~~~Liangyan Gui$^1$~~~Alexander Schwing$^1$\\
${}^1$~University of Illinois at Urbana-Champaign \quad ${}^2$~Waymo}

\maketitle
\begin{abstract}
Forecasting of a representation is important for safe and effective autonomy. For this, panoptic segmentations have been studied as a compelling representation  in recent work. However, recent state-of-the-art on panoptic segmentation forecasting suffers from two  issues: first, individual object instances are treated independently of each other; second, individual object instance forecasts are merged in a heuristic manner. 
To address both issues, we study a new  panoptic segmentation forecasting model that jointly forecasts all object instances in a scene using a transformer model based on `difference attention.' It further refines the predictions by taking depth estimates into account. 
We evaluate the proposed model on the Cityscapes and AIODrive datasets. We find  difference attention to be particularly suitable for forecasting because the difference of quantities like locations enables a model to explicitly reason about velocities and acceleration. Because of this, we attain state-of-the-art on panoptic segmentation forecasting metrics.
\end{abstract}
\section{Introduction}
\label{sec:intro}
Forecasting is needed for safe and effective autonomous systems~\cite{craik1943,llinas2001}. For this reason, forecasting has been studied in many different domains from computer vision and robotics to machine learning. In common across domains is the discussion about what representations are useful for forecasting. Representations which have been studied range from trajectories~\cite{dai2020self, martinez2017human, ehrhardt2020relate,YehCVPR2019,graber2020dynamic} and bounding boxes~\cite{YagiCVPR2018,YaoICRA2019,styles2020multiple,MallaCVPR2020,MakansiICCV2021} to semantic segmentation~\cite{luc2017predicting,rochan2018future,chiu2020segmenting,vsaric2019single,lin2021predictive}, instance segmentation~\cite{luc2018predicting,hu2021apanet,couprie2018joint}, images~\cite{liang2017dual,gao2019disentangling, ye2019compositional} and  recently also panoptic segmentations~\cite{graber2021panoptic,vsaric2021dense}.

Each representation has applications which benefit from their use. We focus on panoptic segmentations as they naturally disentangle 1) objects which change position in an  image due to observer motion; from 2) object instances which change position due to both observer and instance motion. %

\begin{figure}
    \centering
    \includegraphics[width=0.9\linewidth]{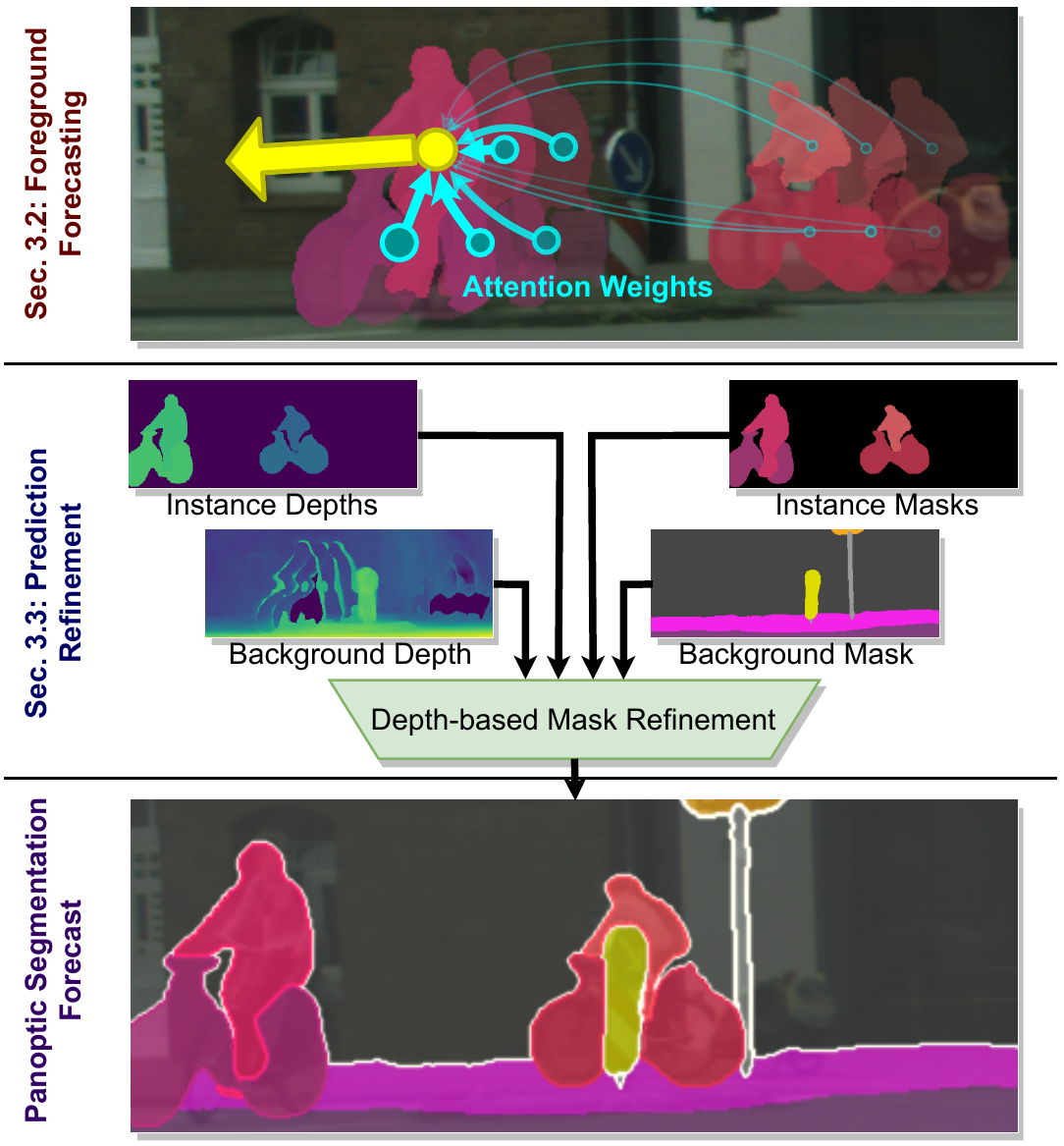}
    \vspace{-0.3cm}
    \caption{Our panoptic segmentation forecasting. We  jointly reason about every instance in a scene to predict instance masks (top), and then reason about the relative depth of foreground and background components (middle) to produce an output (bottom).}
    \label{fig:teaser}
    \vspace{-0.5cm}
\end{figure}

However, the state-of-the-art on panoptic segmentation forecasting~\cite{graber2021panoptic} is challenged by  two key issues. First, foreground predictions of individual instances are made independently of each other. This is suboptimal because the movements of instances are clearly correlated, \eg, when considering traffic patterns like the ones in the Cityscapes dataset~\cite{cordts2016cityscapes}. Second, the method opted for a simple strategy to merge individual object instance segmentation forecasts with the background forecast. Specifically, in~\cite{graber2021panoptic}, object instance segmentation forecasts are always placed in front of the background segmentation forecast. This assumes that no background objects are located closer to the camera than any foreground entity, which is not true in practice. %

In this work, we study a new method to address these two issues: 1) To jointly forecast  object instance segmentations, we develop a modified attention module for transformer models. Specifically, instead of the inner-product attention in classical transformers, we propose ``difference attention.'' This developed difference attention fits tasks like forecasting because it enables reasoning about velocities and acceleration, which is non-trivial with classical inner-product attention (see \cref{fig:teaser} top). 
2) To properly reason about object and background placement, we develop a refinement head  which  %
denoises background depth estimates and compares them against foreground predictions
(see \cref{fig:teaser} middle). 

We assess our method on the challenging Cityscapes~\cite{cordts2016cityscapes} and AIODrive~\cite{Weng2020_AIODrive} datasets. %
We find  difference attention and refinement to provide accurate results (see \cref{fig:teaser} bottom) which yield a new state-of-the-art of $37.6$ PQ for  mid-term forecasting on Cityscapes and $48.5$ PQ on AIODrive. Code to reproduce results is available via \url{https://github.com/cgraber/psf-diffattn}.

\section{Related work}
Forecasting has been studied across communities~\cite{Valassakis2018}.

\noindent\textbf{Forecasting of non-semantic representations.} 
Trajectories are arguably one of the representations for which forecasting has been studied most. Trajectories specify the future position of individual objects, either in 2D or 3D \cite{dai2020self, martinez2017human, ehrhardt2020relate,YehCVPR2019}. 
For example, Hsieh \etal~\cite{hsieh2018learning} disentangle position and pose of multiple moving objects -- but only on synthetic data.
Mittal \etal~\cite{mittal2020just} forecast scene flow for point cloud data using self-supervision to  reduce training data requirements.
Kosiorek~\etal~\cite{kosiorek2018sequential} track instances to forecast their future.
Several works have focused on anticipating future pose and location of specific object types, often people \cite{mangalam2020disentangling, graber2020dynamic}. 
However, arguably, a trajectory forecast provides little  beyond  position, velocity and acceleration.

To obtain  more information, forecasting of future RGB frames has been studied~\cite{liang2017dual,gao2019disentangling, ye2019compositional}. 
Due to the high-dimensional space of the forecasts and because of the ambiguity in the forecasts, results often remain blurry, despite significant recent advances.
For instance, recent work models uncertainty over future frames using, \eg,  latent variables~\cite{walker2016uncertain, ye2019compositional} or treats foreground and background separately~\cite{wu2020future}. %
Moreover, Ye \etal \cite{ye2019compositional} forecast future RGB frames by modeling each foreground object separately.  
Note, all these methods differ from ours in architecture and output: we  forecast a semantic representation.

Closer to our work is AgentFormer~\cite{yuan2021agentformer}. It also uses transformers to forecast and introduces an identity encoding via agent-aware attention. Our work differs in that we predict panoptic segmentations while they predict birds-eye-view locations. Additionally, we develop difference attention and auxiliary losses which we find to aid forecasting.

\noindent\textbf{Forecasting semantic segmentations.} 
Recently,  methods have been studied to estimate semantic segmentations for future, unobserved frames.  
Luc \etal~\cite{luc2017predicting} use a deep-net to estimate a future semantic segmentation given the current RGB frame and its semantics as input.  Nabavi \etal~\cite{rochan2018future} use recurrent models with semantic maps as input. 
Chiu \etal~\cite{chiu2020segmenting} further use a teacher net to provide an additional supervision  during training. \v{S}ari\'{c} \etal~\cite{vsaric2019single} use learnable deformations to help forecast future semantics given the observed frames. Lin \etal~\cite{lin2021predictive} design an autoencoder which 1) compresses input feature pyramids into a low-resolution predictive feature map, 2) predicts this representation for a future frame, and 3) expands it back into a feature pyramid for decoding.
However, importantly, these methods do not explicitly consider dynamics of the scene.

While Jin~\etal~\cite{jin2017predicting} jointly predict flow and future semantic segmentations, recent work~\cite{saric2020warp} explicitly warps deep features to obtain a future semantic segmentation.
Similarly, Terwilliger \etal~\cite{terwilliger2019recurrent} use a long-short-term-memory (LSTM) module to estimate a flow field which is then used to warp the semantic output of a given input frame.
However, by warping in output space,  their model has a limited ability to cope with occlusions.
While flow improves the modeling of the dynamic world, these methods only consider the dynamics at the pixel-level. Instead, we model dynamics at the object level. %

Recent methods~\cite{qi20193d, vora2018future, xu2018structure, hoyer2019short} estimate future semantic segmentations by reasoning about shape, egomotion, and foreground motion separately.
However, none of these methods reason explicitly about individual instances, while our method yields a full future panoptic segmentation forecast, \ie, a prediction for every instance.

\begin{figure*}[t]
    \centering
    \includegraphics[width=\textwidth]{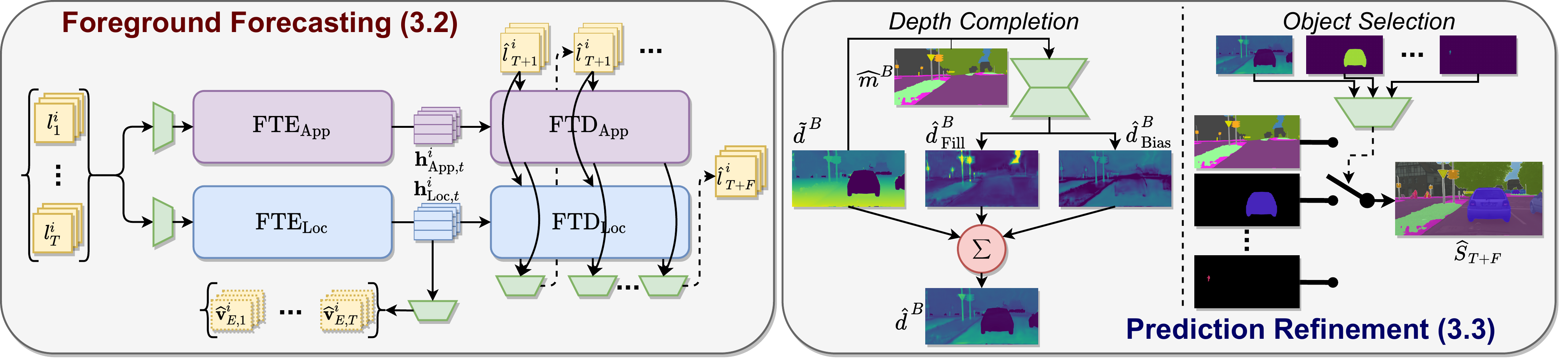}
    \vspace{-0.7cm}
    \caption{\textbf{Method overview}. The foreground forecasting component (\cref{foreground}) predicts future location and appearance jointly for all instances using our newly introduced difference attention transformer. This is followed by a prediction refinement stage (\cref{refine}), which first completes/denoises input reprojected depths and then uses these along with predicted foreground instance depths to select which object is closest to the camera.}
    \label{fig:model}
    \vspace{-0.4cm}
\end{figure*}

\noindent\textbf{Forecasting future instance segmentations.} 
Recent methods which forecast an instance segmentation use a conv net or an LSTM module to regress to the deep features which correspond to the future instance segmentation \cite{luc2018predicting,hu2021apanet}.
For example, Couprie \etal \cite{couprie2018joint} use a conv net to forecast future instance contours together with an instance-wise semantic segmentation to estimate future instance segmentation. 
However, their method only estimates foreground and not background semantics.

Unlike these works, we predict both instance segmentation masks for foreground objects and background semantics for future time steps.

\noindent\textbf{Forecasting panoptic segmentations.} 
In recent years, panoptic segmentation has become a popular scene understanding task \cite{Cheng2020panoptic-deeplab,cheng2021maskformer,li2021fully,wang2021max,cheng2021mask2former}.
Very recently~\cite{graber2021panoptic,vsaric2021dense}, it has been proposed 
as a useful representation for forecasting because it naturally disentangles 1) objects which move in an image just because of observer motion; from 2) object instances which move due to both observer and instance motion.

The state-of-the-art~\cite{graber2021panoptic}  forecasts the future position of individual object instances independently of each other via an encoder-decoder architecture which is executed  separately for every object instance. Moreover,  the obtained instance forecasts are combined in a heuristic manner by simply pasting objects in front of background without considering depth information of background objects. %

In contrast, we propose a method for panoptic segmentation forecasting which jointly forecasts all detected object instances at once via a tailored transformer attention. This helps to benefit from correlations between instances. Moreover, we study how to %
combine the individual forecasts in a differentiable way. We discuss our method next.

\section{Method}
In this section, we describe our method for joint forecasting of multiple object instances with the developed difference attention. %
We start by formalizing the forecasting task and by providing an overview of our approach. Next, we describe the developed difference attention  (\cref{transformer}). We use this in our %
foreground forecasting module, which models interactions between individual instances (\cref{foreground}). Finally, we present the refinement head, which refines the initial foreground instance predictions by considering the background predictions and the depth (\cref{refine}). An overview of our approach is presented in \cref{fig:model}.

\noindent\textbf{Forecasting task.} Given  $T$ RGB images $I_1, \dots, I_T$ of height $H$ and width $W$, panoptic segmentation forecasting aims to predict the panoptic segmentation $\widehat{S}_{T+F}$  corresponding to an unobserved future frame $I_{T+F}$ at a fixed number of timesteps $F$ from the last observation recorded at time $T$. Each pixel in $\widehat{S}_{T+F}$ is assigned a class $c \in \{1, \dots, C\}$ and an instance ID.  In addition to these inputs, we assume access to camera poses $o_1, \dots, o_T$ and depth maps $d_1, \dots, d_T$ for all input frames. 
We study the use of both camera poses from odometry sensors, and camera poses estimated using visual SLAM \cite{ORBSLAM3_2020}. 
Following \cite{graber2021panoptic}, we obtain depth maps from input stereo image pairs \cite{gu2020cascade}. %

\noindent\textbf{Overview.}
To address forecasting, we follow the paradigm introduced by Graber \etal~\cite{graber2021panoptic}. Specifically, they divide the task into two  components: 1) the foreground component, which focuses on `things' object instances annotated within the dataset; and 2) the background component, which focuses on all annotated `stuff' object classes. These two components are modeled differently because the causes for the displacement of the corresponding objects in the image plane differ. Specifically, background objects such as buildings and poles shift due to camera motion, while foreground objects like cars and pedestrians move due to both camera motion as well as their own individual motion.

For a fair evaluation, we utilize the same background model as Graber \etal~\cite{graber2021panoptic}, who lift background semantics into a 3D point cloud using the estimated input depth, transform the depth based on the target frame camera information, project to the image plane, and refine the projected semantics using a semantic segmentation model. See Appendix~\ref{app:bgmodel} for more details. 

However, the approach developed by Graber \etal~\cite{graber2021panoptic} has two primary drawbacks which we correct in this work: 

First, their approach to forecast the foreground components of the scene uses an RNN-based encoder-decoder model which models the trajectory of each instance independently of all other instances. This is sub-optimal: in many cases, the movement of individual entities is correlated, \eg, due to the flow of traffic. 
To enable modeling of this correlation, we develop a difference attention module which we detail in \cref{transformer}. It is particularly suitable for forecasting because of its innate ability to reason about the velocities of inputs.
We use this difference attention transformer to jointly reason about the future trajectories of all entities in a scene, which we detail in \cref{foreground}.

Second, Graber \etal~\cite{graber2021panoptic} combine foreground and background predictions by ``stacking'' all predicted foreground instances on top of the predicted background. This approach assumes that no background objects are located closer to the camera than any foreground entity, which is not true in practice. Hence, in \cref{refine}, we introduce our model to combine foreground and background predictions  in a per-pixel fashion by reasoning about their  depths.

\subsection{Difference Attention for Transformers}
\label{transformer}

To better address forecasting, we develop a difference attention module for transformers. 
We find this difference attention to be particularly suitable for forecasting because the difference of quantities like locations enables a model to easily reason about velocities and acceleration. In contrast, classical transformer attention is based on inner products which do not naturally encode these quantities.

Formally, the difference attention module operates on two $d$-dimensional inputs $\mathbf{X}_\text{self} \in \mathbb{R}^{M_1 \times d}$ and $\mathbf{X}_\text{other} \in \mathbb{R}^{M_2 \times d}$ of lengths $M_1$ and $M_2$, respectively, reasons about the differences between these inputs, and outputs representation $\mathbf{Y} \in \mathbb{R}^{M_1 \times d}$ which encodes these differences. For this, we first compute entity scores 
\begin{equation}
\mathbf{Z} = \mathbf{Q} \mathbf{K}_R^T - \mathbf{1}_{M_1 \times 1}\text{diag}\left(\mathbf{K}_B \mathbf{K}_R^T \right)^T,
\end{equation}
where $\mathbf{1}_{M_1 \times 1}$ is the $M_1 \times 1$ matrix filled with ones. $\mathbf{Q}$ is computed from $\mathbf{X}_\text{self}$ and $\mathbf{K}_B$ and $\mathbf{K}_R$ are computed from $\mathbf{X}_\text{other}$ with MLPs, \ie, $\mathbf{Q} = f_Q(\mathbf{X}_\text{self})$, $\mathbf{K}_R = f_{K_R}(\mathbf{X}_\text{other})$, and $\mathbf{K}_B = f_{K_B}(\mathbf{X}_\text{other})$. Intuitively, this operation allows the entity score computation to be a function of the difference between the two inputs $\mathbf{X}_\text{self}$ and $\mathbf{X}_\text{other}$. This is useful for forecasting, as the offset of input locations and their change over time is necessary to understand motion.%

Given these entity scores $\mathbf{Z}$, we compute the final attended representation $\mathbf{Y}$ which corresponds to $\mathbf{X}_\text{self}$ via
\begin{equation}
   \mathbf{Y} = \text{softmax}\left(\mathbf{Z} / \sqrt{d} \right) \mathbf{V}_O - \mathbf{V}_S,
\end{equation}
where $\mathbf{V}_O = f_{V_O}(\mathbf{X}_\text{other})$ and $\mathbf{V}_S = f_{V_S}(\mathbf{X}_\text{self})$. Intuitively, this enables the final output $\mathbf{Y}$ to encode the differences between the two inputs $\mathbf{X}_\text{self}$ and $\mathbf{X}_\text{other}$. This is again suitable for forecasting, as it enables representations to encode  the velocity of an instance, which is critical for reasoning about future motion. 
We now discuss how we use this difference attention  for foreground forecasting.

\subsection{Foreground Forecasting}
\label{foreground}

Our forecasting model is tasked with predicting a panoptic segmentation $\widehat{S}_{T+F}$ for time ${T+F}$. This is done by forecasting representations for the $N$ instances in the scene, followed by a final refinement. We represent each instance at all times during forecasting using three components $l_t^i \coloneqq \{\mathbf{x}_t^i, \mathbf{r}_t^i, p_t^i\}$: a 5-dimensional vector $\mathbf{x}_t^i \coloneqq [x_0, y_0, x_1, y_1, d]$ representing the upper-left and lower-right corners of the bounding box enclosing instance $i$ as well as the estimated distance of the instance from the camera at time $t$, a feature tensor $\mathbf{r}_t^i \in \mathbb{R}^{256 \times 14 \times 14}$ representing the visual appearance of the instance at time $t$, and a binary value $p_t^i \in \{0, 1\}$ which indicates whether instance $i$ is present in frame $I_t$. Additionally, given background prediction logits $\widehat{m}^B$ and background reprojected depths $\tilde{d}^B$, the final output of the forecasting model is 
\begin{equation}
    \widehat{S}_{T+F} = \text{Ref}(\text{FD}(\text{FE}(\{l_t^i, c^i, o_t\}_{1:T}^{1:N}), \{o_t\}_{T+1:F}), \widehat{m}^B, \tilde{d}^B). 
\end{equation}
Here, the forecasting encoder FE operates on input representations $l_t^i$, classes $c^i$, and odometry $o_t$ $\forall i \in \{1, \dots, N\}, t \in \{1, \dots, T\}$ and computes embeddings $\mathbf{h}^i_{\text{Loc}, t}$ and $\mathbf{h}^i_{\text{App}, t}$ which encode locations and appearances, respectively. The forecasting decoder FD processes these embeddings to autoregressively compute embeddings $\tilde{\mathbf{h}}^i_{\text{Loc}, t}$ and $\tilde{\mathbf{h}}^i_{\text{App}, t}$, which are used to produce outputs $\widehat{l}_t^i$. These outputs are subsequently combined with background semantics $\widehat{m}^B$ and depths $\tilde{d}^B$ using refinement model Ref to produce the final panoptic segmentation  $\widehat{S}_{T+F}$.
We discuss the encoder and  decoder  which use  difference attention  next, and we detail refinement  in \cref{refine}.

\noindent\textbf{Forecasting Transformer Encoder.} 
The encoder $\text{FE}$ produces two embeddings for every instance $i$ at every time $t$: the first, $\mathbf{h}_{\text{Loc},t}^{i} \in \mathbb{R}^{d_{e}}$ where $d_e$ is the size of the embedding, contains information about its location as well as its observed motion; the second, $\mathbf{h}_{\text{App},t}^{i} \in \mathbb{R}^{256 \times 14 \times 14}$, contains information about its appearance. 
These are obtained using two newly developed forecasting transformer encoders. %
The use of transformers for this task permits to jointly reason about every instance both as a function of time and as a function of the other instances present in the scene.

The first transformer encoder produces in parallel $\forall i,t$ the location encoding  
\begin{equation}
\{\mathbf{h}_{\text{Loc},t}^{i}\}_{1:T}^{1:N} = \text{FE}_\text{Loc}(\{l_t^i, c^i, o_t\}_{1:T}^{1:N}).
\label{eq:hloc}
\end{equation}
For this, it uses all input instances at every point in time, \ie,  $\{l_i^t\}_{1:T}^{1:N}$, as well as classes $c^i$ and odometry $o_t$. Different from classical transformer encoders, $\text{FE}_\text{Loc}$ is trained via auxiliary losses to natively reason about both the velocity of each instance across time as well as the motion of each instance relative to each other. Hence, the embedding $\mathbf{h}_{\text{Loc},t}^{i}$ is trained to encode information about the velocity, which we show improves the ability of the decoder to anticipate the instances' future motion. 

The second transformer encoder, which produces the appearance encoding  \begin{equation}
\mathbf{h}_{\text{App},t}^{i} = \text{FE}_\text{App}(\{l_t^i, c^i, o_t\}_{1:T}^{1:N}),
\label{eq:happ}
\end{equation}
maintains the spatial structure of the input appearance features. This is beneficial for predicting a  spatial output. 

Both the location and the appearance components of the forecasting transformer encoder are comprised of the same general structure: first, a feature representation for every instance is produced as a function of its location, its appearance, its object class, the current camera motion, and the current time. Second, these feature representations are processed using our customized transformer encoders $\text{FTE}_\text{Loc}$ and $\text{FTE}_\text{App}$. Letting $\beta \in \{\text{Loc}, \text{App}\}$ denote the modules for the location encoder and the appearance encoder, respectively, this is formally described as
\begin{eqnarray}
&&\hspace{-1cm}\{\mathbf{h}_{\beta,t}^{i}\}_{1:T}^{1:N} = \text{FE}_\beta(\{l_t^i, c^i, o_t\}_{1:T}^{1:N}) \label{eq:one}\\
&&\hspace{-0.3cm}\Leftrightarrow \left\{\begin{array}{rll}
    \bar{\textbf{x}}_{\beta,t}^i &= f_\beta(l_t^i, c_i, o_t, t)&\forall i,t \\
    \{\mathbf{h}_{\beta,t}^{i}\}_{1:T}^{1:N} &= \text{FTE}_\beta(\{\bar{\textbf{x}}_{\beta,t}^i\}_{1:T}^{1:N})
\end{array}\right., \label{eq:two}
\end{eqnarray}
where $f_\text{Loc}$ uses multilayer perceptrons and $f_\text{App}$ uses convolutional nets which are described fully in Appx.~\ref{app:locappfeat}. Note, depending on $\beta$, Eq.~\eqref{eq:one} refers to either Eq.~\eqref{eq:hloc} or Eq.~\eqref{eq:happ}. They perform the computations given in Eq.~\eqref{eq:two}. All features $\{\bar{\textbf{x}}_{\beta,t}^i\}$ are used as input into the transformer $\text{FTE}_\beta$.

For $\text{FTE}_\text{Loc}$, all self-attention modules use the difference attention formulation introduced in \cref{transformer}. This design facilitates the ability of the model to reason about the velocity of the entities, which can be represented by differences in input embeddings which correspond to the same instance at different points in time, as well as the relative offsets between different entities. We find that the use of this form of attention leads to improved forecasting results.

The appearance transformer encoder $\text{FTE}_\text{App}$ is built using convolutional transformers. %
Specifically, it consists of a transformer whose linear projections have been replaced with convolutional layers. This enables a spatially meaningful representation at all stages during encoding. 

For more about attention computation see Appendix~\ref{app:agentawareattention}.

\noindent\textbf{Forecasting Transformer Decoder.} 
The decoder utilizes the representations produced by the encoder to predict the future location $\widehat{\textbf{x}}_i^t$, the future appearance $\widehat{\textbf{r}}_i^t$, and the future presence $\widehat{p}_i^t$ of each object $i$ for future time steps $t\in \{T+1,\dots,T+F\}$. Predictions are computed autoregressively, starting with the most recent input locations $\widehat{\textbf{x}}_i^T := \mathbf{x}_i^T$ and appearance features $\widehat{\textbf{r}}_i^T := \mathbf{r}_i^T$. 

For future time step $t \in \{T+1, \dots, T+F\}$, both the location decoder $\text{FD}_\text{Loc}$ and the appearance decoder $\text{FD}_\text{App}$ take the following structure, with $\beta \in \{\text{Loc}, \text{App}\}$:
\begin{eqnarray}
&&\hspace{-1cm}\{\tilde{\mathbf{h}}_{\beta,t}^{i}\}^{1:N} = \text{FD}_\beta(\{\widehat{l}_t^i\}^{1:N}_{T:t-1}, \{c^i\}^{1:N}, \{o_t\}_{T+1:t},\{\mathbf{h}_{\beta,t}^{i}\}_{1:T}^{1:N}) \nonumber\\
&&\hspace{-0.3cm}\Leftrightarrow \left\{\begin{array}{rll}
    \tilde{\textbf{x}}_{\beta,t}^i &= \tilde{f}_\beta(\widehat{l}_t^i, c^i, o_t, t)&\forall i,t \\
    \{\tilde{\mathbf{h}}_{\beta,t}^{i}\}_{T+1:T+F}^{1:N} &= \text{FTD}_\beta(\{\tilde{\textbf{x}}_{\beta,t}^i\}_{T+1:T+F}^{1:N})
\end{array}\right..\nonumber
\end{eqnarray}
Similar to their corresponding encoder modules, the location transformer decoder $\text{FTD}_\text{Loc}$ uses difference attention, the appearance transformer decoder $\text{FTD}_\text{App}$ is a convolutional transformer, and both utilize agent-aware attention.

Final location, appearance, and presence predictions are obtained from the embeddings produced at each time via
\begin{align}
    \widehat{\mathbf{x}}_t^i &= f_\text{LocOut}(\tilde{\mathbf{h}}_{\text{Loc},t}^i) + \widehat{\mathbf{x}}_{t-1}^i, \\
    \widehat{p}_t^i &= f_\text{POut}(\tilde{\mathbf{h}}_{\text{Loc},t}^i), \\
    \widehat{\mathbf{r}}_t^i &= f_\text{AppOut}(\tilde{\mathbf{h}}_{\text{App},t}^i),
\end{align}
where $f_\text{LocOut}$ and $f_\text{POut}$ are multilayer perceptrons and $f_\text{AppOut}$ is a convolutional network.

\noindent\textbf{Training.}
The foreground model is trained by providing it with input location and appearance features, predicting the future states of each of these, and regressing against pseudo-ground-truth future locations $\mathbf{x}_t^{*i}$, appearance features $\mathbf{r}_t^{*i}$, and presences $p_t^{*i}$ which are obtained by running instance detection and tracking on future frames. We formally specify the losses in Appendix~\ref{app:foregroundforecastinglosses}.

In addition,  we train the forecasting location encoder to estimate the velocity $\widehat{\textbf{v}}_{\text{E},t}^i$ of each instance via
\begin{equation}
    \widehat{\textbf{v}}_{\text{E},t}^{i} = f_\text{vel}(\textbf{h}_{\text{Loc},t}^{i}),
    \label{eq:enc_vel}
\end{equation}
where $f_\text{vel}$ is a multilayer perceptron. This auxiliary prediction task requires the encoder to include information about the motion of each instance within the representation it produces. We find  this to lead to better forecasting results.

\subsection{Prediction Refinement}
\label{refine}
To address the aforementioned second shortcoming of \cite{graber2021panoptic}, we develop a   refinement  which combines foreground and background predictions as a function of their estimated depth. This  allows foreground instances to be placed behind background objects, which yields more natural predictions.

While this would be easy if the depth signal was reliable, 
the only depth signal we have for the background is the depth of the reprojected points that are used as input for the background prediction model. These depths are both noisy and incomplete, \ie, not every location will correspond to a reprojected point from an earlier frame. Hence, the refinement model has two primary jobs: first, it needs to complete as well as denoise the input depth; second, it needs to   select which object is closest based on these depths as well as the depths of foreground instances. %

Formally, the refinement head is provided with predicted foreground locations\footnote{For readability, we drop subscript $T\!+\!F$  for  predictions in this section.} $\widehat{\mathbf{x}}^i$, appearances $\widehat{\textbf{r}}^i$, and presences $\widehat{p}^i$ for $N$ instances. Given these components, if $p^i = 0$, then instance $i$ is discarded, as the model  anticipates that the object is not in frame $I_{T+F}$ due to occlusions or leaving the scene; otherwise, the prediction mask $\widehat{m}^i$ is obtained via
\begin{equation}
    \widehat{m}^i = \text{MaskOut}(\widehat{\mathbf{x}}^i, \widehat{\textbf{r}}^i),
\end{equation}
where \text{MaskOut} predicts a fixed-size mask using MaskRCNN's mask head and then pastes it into the location specified by $\widehat{\mathbf{x}}^i$.
The prediction head additionally uses estimated instance depths $\{\widehat{d}^i\}^{1:N}$, predicted background semantic logits $\widehat{m}^B \in \mathbb{R}^{H \times W \times C_\text{BG}}$, where $C_{BG}$ is the number of background classes, the reprojected background depths $\tilde{d}^B \in \mathbb{R}^{H \times W}$, and a binary mask $Q \in \{0, 1\}^{H \times W}$ which indicates for each pixel whether or not we have an input background depth. It  outputs an object selection map $\widehat{P} \in \{0, \dots, N\}^{H \times W}$ which specifies, for every pixel, whether the background is in front (represented by value $0$) or one of the instances is in front (represented by values $1$ through $N$). %
We get the final panoptic segmentation  via 
\begin{equation}
    \widehat{S}_{T+F}\!=\! \mathbf{1}[\widehat{P} = 0]\argmax(\widehat{m}^B) \!+\! \mathbf{1}[\widehat{P} > 0](\widehat{P} + C_\text{BG}).\nonumber
\end{equation}

The refinement head is composed of two modules: the first produces completed/denoised background depth prediction $\widehat{d}^B$, and the second uses this alongside the foreground instance information to compute the object selection map $\widehat{P}$. We  describe both components next.

\noindent\textbf{Depth completion model.}
We formulate the depth completion model using two outputs. The first, $\widehat{d}_{\text{Fill}}^B \in \mathbb{R}^{H \times W}$, represents an initial estimation of the depths for all input locations which are missing a depth, \ie, where $Q = 0$. The second, $\widehat{d}_{\text{Bias}}^B\in \mathbb{R}^{H \times W}$, represents an offset added to the input depths in order to refine and denoise them. Given these predictions, the output of this module is 
\begin{align}
    \widehat{d}^B &= Q \tilde{d}^B + (1-Q) \widehat{d}_{\text{Fill}}^B + \widehat{d}_{\text{Bias}}^B,
    \label{eq:bgd}
\end{align}
where $\widehat{d}_{\text{Fill}}^B$ and $\widehat{d}_{\text{Bias}}^B$ are obtained using small convolutional networks specified in Appendix~\ref{app:depthcompletion}.

\noindent\textbf{Object selection model.}
Given the completed/denoised background depth prediction $\widehat{d}^B$, object selection determines for every output pixel which object is closest to the camera. We require that this module be fully differentiable such that gradients computed from its outputs can be propagated through to the depth completion model.

More formally, we compute the aggregate depth tensor $\mathbf{D} \in \mathbb{R}^{H\times W \times (N+1)}$ whose $0$-th channel is the completed background depth $\widehat{d}^B$ and whose $i$-th channel for $i \in \{1, \dots, N\}$ is $\mathbf{1}[\widehat{m}^i \geq 0.5]d^i + \mathbf{1}[\widehat{m}^i < 0.5]d_\text{fgmax}$. Here, $\mathbf{1}[\cdot]$ is the indicator function applied to all spatial locations in $\widehat{m}^i$. Further, $d_\text{fgmax}$ is a large constant. We also construct a value tensor $\mathbf{V} \in \mathbb{R}^{H \times W \times (N+1)}$ whose $n$-th channel is computed by applying a convolutional net to the  background logits $\widehat{m}^B$ and foreground probabilities $\widehat{m}^i$. The final prediction is computed via
\begin{equation}
    \tilde{P}_{i,j} = \text{softmax}\left(-\mathbf{D}_{i,j}\right)\circ \mathbf{V}_{i,j},
\end{equation}
where $\tilde{P} \in \mathbb{R}^{H \times W \times (N+1)}$ are object selection scores for each pixel, $\widehat{P}_{i,j} = \argmax{\tilde{P}_{i,j}}$, and $\circ$ is the Hadamard product. Specifically, for each pixel location $(i,j)$, we use the softmax function to determine the smallest depth and then multiply by the value vector to attain correct scaling of the output probabilities.

\noindent\textbf{Training.}
For training, we  compute the input instance masks $\widehat{m}_{T+F}^i$ using the pseudo-ground-truth locations $\mathbf{x}_t^{*i}$. We then obtain completed background depths and compute final object selection scores $\tilde{P}_{T+F}$. This is compared to the ground-truth object selection $P_{T+F}^*$ using cross-entropy. We additionally apply a squared norm loss to the predicted depth bias $\widehat{d}_\text{Bias}^B$ such that the model is encouraged to trust the input depths where possible. Note that we do not supervise the depth completion model to predict globally accurate depths. Instead, we only require that the completed background depths have the correct relative value compared to the foreground depths, \ie,  the predicted depths lead to selecting either the foreground or the background correctly per-pixel.

\begin{table*}[t]
    \vspace{-0.2cm}
    \footnotesize
      \centering
      \resizebox{1.0\textwidth}{!}{
      \begin{tabular}{l ccc ccc ccc c ccc ccc ccc}
    \toprule
    & \multicolumn{9}{c}{Short term: $\Delta t = 3$} && \multicolumn{9}{c}{Mid term: $\Delta t = 9$} \\
    & \multicolumn{3}{c}{All} & \multicolumn{3}{c}{Things} & \multicolumn{3}{c}{Stuff}
    && \multicolumn{3}{c}{All} & \multicolumn{3}{c}{Things} & \multicolumn{3}{c}{Stuff} \\
    & PQ & SQ & RQ & PQ & SQ & RQ & PQ & SQ & RQ && PQ & SQ & RQ & PQ & SQ & RQ & PQ & SQ & RQ\\
    \midrule
    Panoptic Deeplab (Oracle)$\dagger$ & $60.3$ & $81.5$ & $72.9$    & $51.1$ & $80.5$ & $63.5$    & $67.0$ & $82.3$ & $79.7$         && $60.3$ & $81.5$ & $72.9$ &$51.1$ & $80.5$ & $63.5$    & $67.0$ & $82.3$ & $79.7$   \\
    \midrule
   Panoptic Deeplab (Last seen frame)  & $32.7$ & $71.3$ & $42.7$      & $22.1$ & $68.4$ & $30.8$ & $40.4$  & $73.3$ & $51.4$  && $22.4$ & $68.5$ & $30.4$ & $10.7$ & $65.1$ & $16.0$ & $31.0$ & $71.0$ & $40.9$ \\
    Flow & $41.4$ & $73.4$ & $53.4$ & $30.6$ & $70.6$ & $42.0$ & $49.3$ & $75.4$ & $61.8$ && $25.9$ & $69.5$ & $34.6$ & $13.4$ & $67.1$ & $19.3$ & $35.0$ & $71.3$ & $45.7$\\
    Hybrid \cite{terwilliger2019recurrent} (bg) and \cite{luc2018predicting} (fg)  &  $43.2$ & $74.1$ & $55.1$ & $35.9$ & $72.4$ & $48.3$ & $48.5$ & $75.3$ & $60.1$ & & $29.7$ & $69.1$ & $39.4$ & $19.7$ &  $66.8$ &  $28.0$ & $37.0$ &  $70.8$ &  $47.7$\\
    IndRNN-Stack~\cite{graber2021panoptic}  & $49.0$ & $74.9$ & $63.3$ & $40.1$ & $72.5$ & $54.6$ & $55.5$ & $76.7$ & $69.5$       & & $36.3$ & $71.3$ & $47.8$ & $25.9$ & $\textbf{69.0}$ & $36.2$ & $43.9$ & $72.9$ & $56.2$\\
    \textbf{Ours} & $\textbf{50.2}$ & $\textbf{75.7}$ & $\textbf{64.3}$ & $\textbf{42.4}$ & $\textbf{74.2}$ & $\textbf{56.5}$ & $\textbf{55.9}$ & $\textbf{76.8}$ & $\textbf{70.0}$ & & $\textbf{37.6}$ & $\textbf{71.4}$ & $\textbf{49.5}$ & $\textbf{28.6}$ & $\textbf{69.0}$ & $\textbf{40.1}$ & $\textbf{44.1}$ & $\textbf{73.2}$ & $\textbf{56.4}$ \\
    \bottomrule
      \end{tabular}}
      \vspace{-0.3cm}
      \caption{\textbf{Panoptic segmentation forecasting evaluated on the Cityscapes validation set}. 
      $\dagger$ has access to the RGB frame at time $T+F$. Higher is better for all metrics.}
      \label{tab:panoptic}
      \vspace{-0.3cm}
 \end{table*}
 
\begin{figure*}[t]
\centering
\footnotesize
\setlength\tabcolsep{0.5pt}
\renewcommand{\arraystretch}{0.5}
\begin{tabular}{cccc}
\rotatebox{90}{\hspace{0.5cm}IndRNN-Stack} & \includegraphics[width=0.32\textwidth]{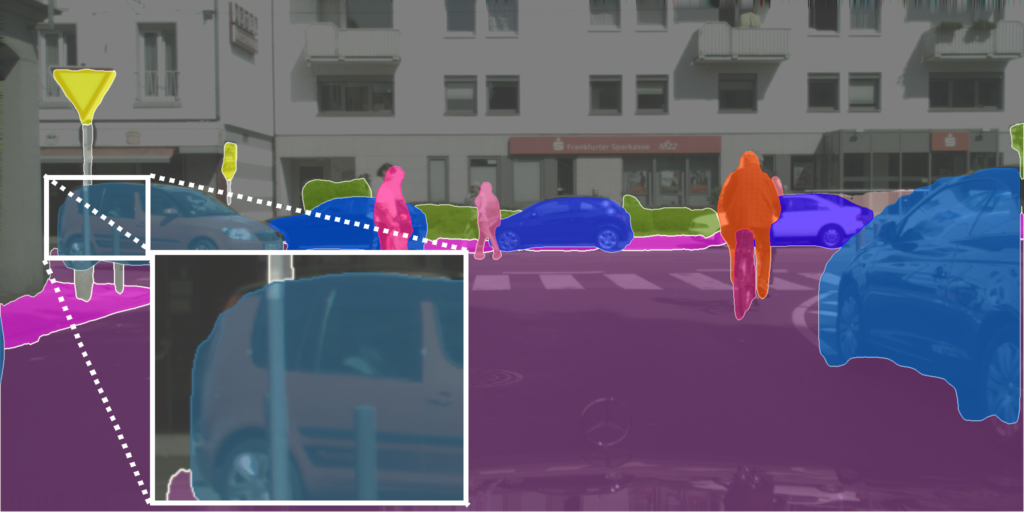}& \includegraphics[width=0.32\textwidth]{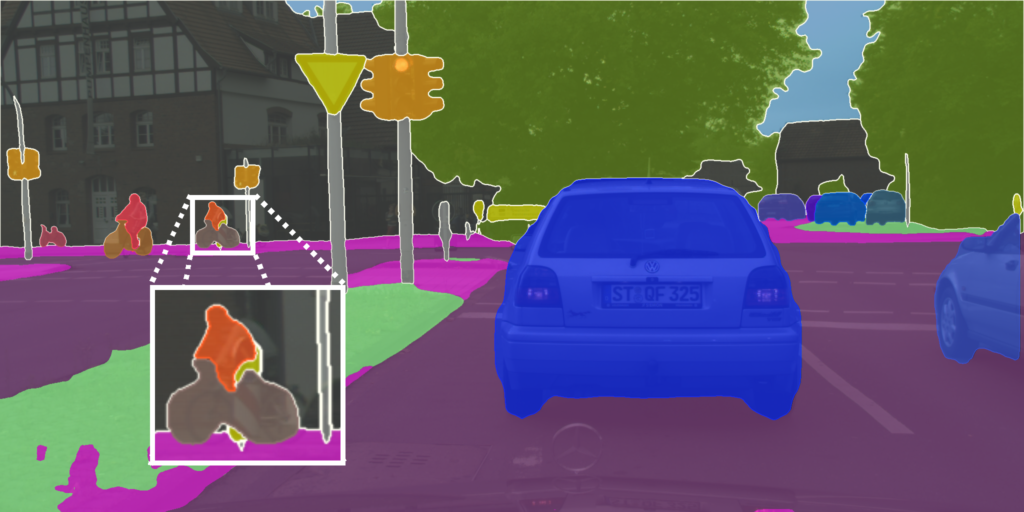} & \includegraphics[width=0.32\textwidth]{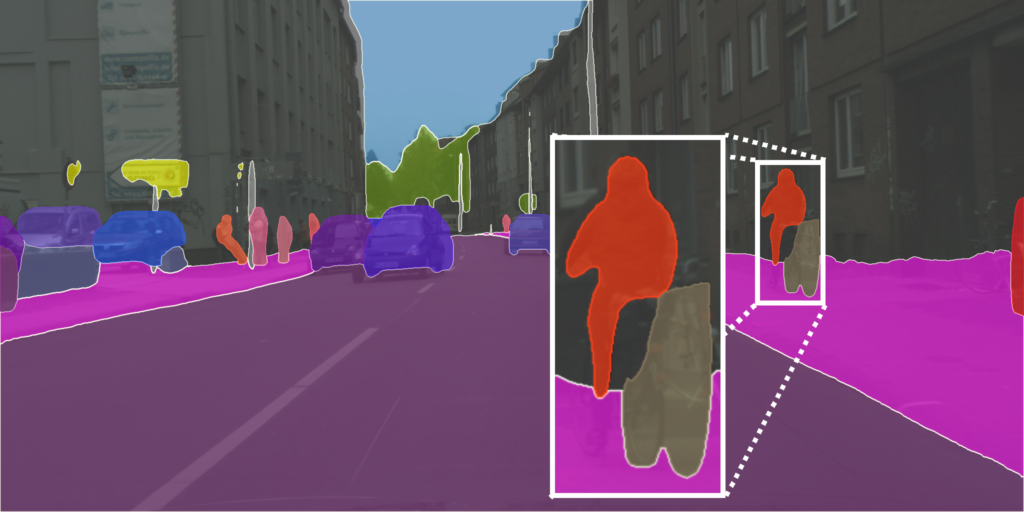} \\ 
\rotatebox{90}{\hspace{1.0cm}\textbf{Ours}} & \includegraphics[width=0.32\textwidth]{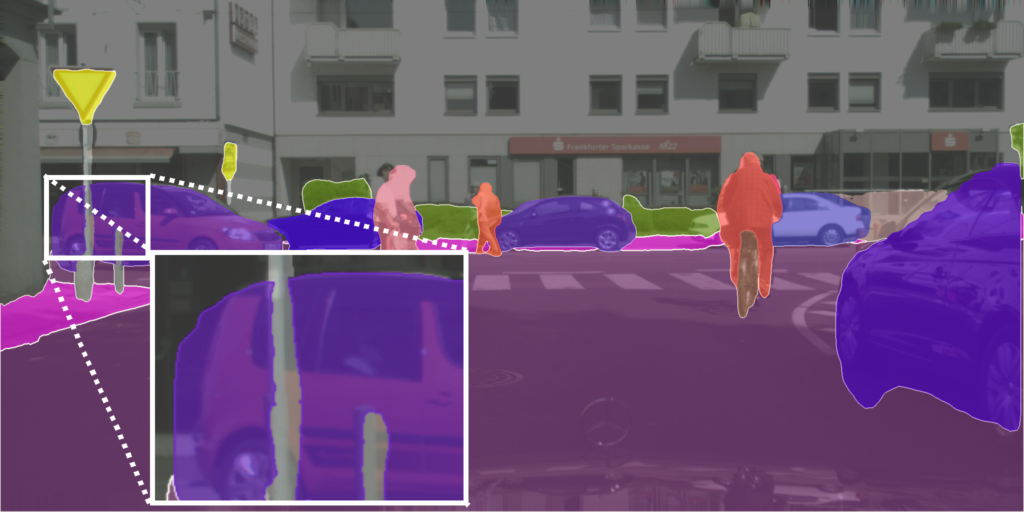}& \includegraphics[width=0.32\textwidth]{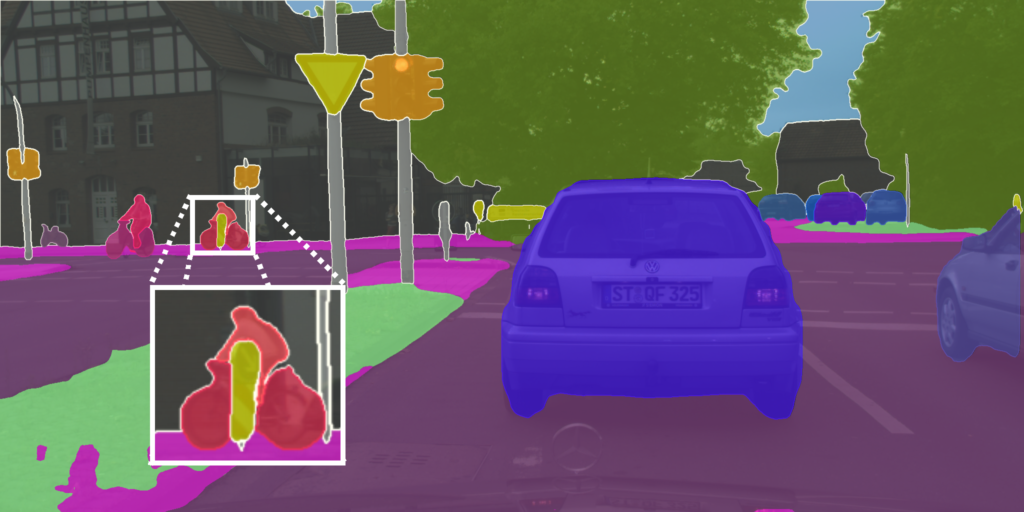} & \includegraphics[width=0.32\textwidth]{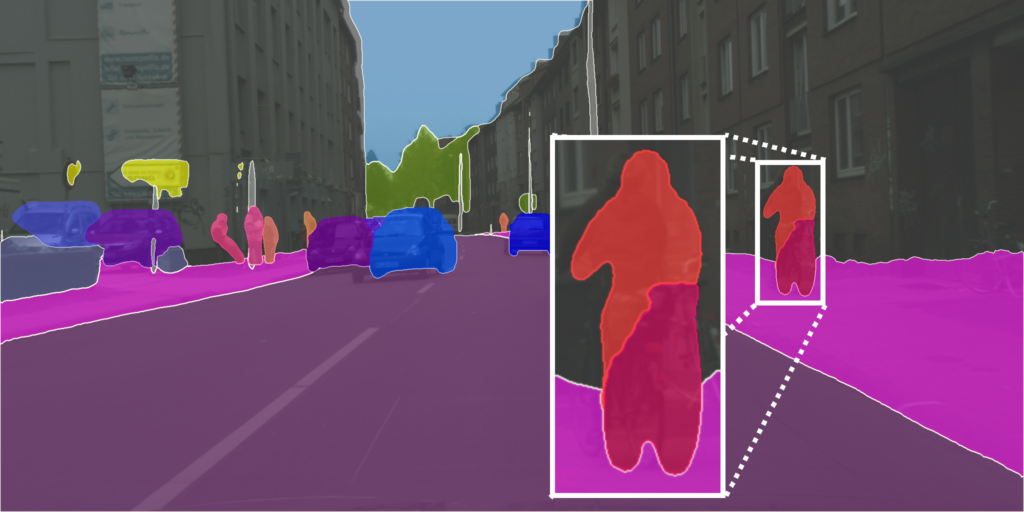}
\end{tabular}
\vspace{-0.3cm}
\caption{Mid-term panoptic segmentation forecasts on Cityscapes. Unlike IndRNN-Stack, our approach is able to properly place foreground instances behind background objects (left two columns). Additionally, our approach models interactions between objects, leading to additional improvements (right column).}
\label{fig:panoptic_viz}
\vspace{-0.6cm}
\end{figure*}

\section{Experiments}
We demonstrate that  the proposed difference attention and refinement  lead to a new state-of-the-art for panoptic segmentation forecasting. We additionally show the contribution each component makes to the final improvement via ablations.  In addition, we demonstrate how these improvements carry over to related dense forecasting tasks.
Following prior work~\cite{graber2021panoptic}, we test our  forecasting model on the Cityscapes dataset~\cite{cordts2016cityscapes}. We additionally run experiments on the recently-introduced AIODrive dataset~\cite{Weng2020_AIODrive}.

\subsection{Cityscapes}
\noindent{\textbf{Data.}}
The Cityscapes dataset contains 5,000 sequences of 30 frames each, where ground-truth panoptic segmentations are provided for the $20$th frame of each sequence. Here, we evaluate our forecasting model on panoptic segmentation forecasting. Additional results for instance segmentation and semantic segmentation forecasting can be found in \cref{app:instseg} and \cref{app:semseg}. 
We consider two types of forecasting: short-term and  mid-term forecasting, each looking 3 and 9 frames into the future respectively. In both cases, we take every third frame as input to our model, hence matching the methods used in prior work \cite{graber2021panoptic,luc2018predicting, luc2017predicting, saric2020warp}.

\noindent{\textbf{Metrics.}}
Following prior work~\cite{graber2021panoptic}, we consider three metrics: \emph{segmentation quality} (SQ), \emph{recognition quality} (RQ), and \emph{panoptic quality} (PQ). First, we match predicted and target segments, where true positive matches require the intersection over union (IoU) of the two segments to be at least 0.5. SQ corresponds to the average IoU of true matched positive segments. RQ corresponds to the F1 score computed over matches. Finally, PQ is the product of SQ and RQ. These metrics are computed for each individual class and then averaged over all classes.

\noindent{\textbf{Baselines.}}
We compare against the baselines introduced in \cite{graber2021panoptic}. \emph{Panoptic Deeplab (Oracle)} applies the Panoptic Deeplab model \cite{Cheng2020panoptic-deeplab} on the target frame, and represents an upper bound on performance due to its access to oracle future information. \emph{Panoptic Deeplab (Last Seen Frame)}  applies this model to the most recently observed frame, which represents a model assuming no camera or instance motion. \emph{Flow} computes optical flow \cite{ilg2017flownet} from the last two observed frames and then uses it to warp the panoptic segmentation obtained from the last observed frame. \emph{Hybrid Semantic/Instance Forecasting} fuses a semantic segmentation forecast \cite{terwilliger2019recurrent} with an instance segmentation forecast \cite{luc2018predicting} to create a panoptic segmentation for the target frame. Finally, IndRNN-Stack is the model introduced by Graber \etal~\cite{graber2021panoptic} which forecasts individual instances using an RNN encoder-decoder model and stacks all foreground components on top of all background components.
 
\noindent{\textbf{Results.}}
The results for all models on the panoptic segmentation forecasting task are presented in \cref{tab:panoptic}. The proposed approach achieves state-of-the-art across both short- and mid-term settings on all metrics when compared to methods which don't access future information. %

\cref{fig:panoptic_viz} presents a visual comparison. IndRNN-Stack is not capable of placing foreground instances behind background objects, which leads to missing segmentations such as  poles in the left column and the street sign in the middle column. Our approach properly reasons about the depth of these objects and places the poles in front of the car and the street sign in front of the cyclist. Additionally, since IndRNN-Stack predicts instances independently, it can make trivial errors such as predicting a cyclist floating away from their bicycle (right column). Our approach, which models interactions among  instances and can reason about the fact that cyclists should always move with their bicycles, does not make this error. Additional visualizations comparing these models are presented in Appendix~\ref{app:additionalviz}.

\begin{figure}[t]
\includegraphics[width=0.49\columnwidth]{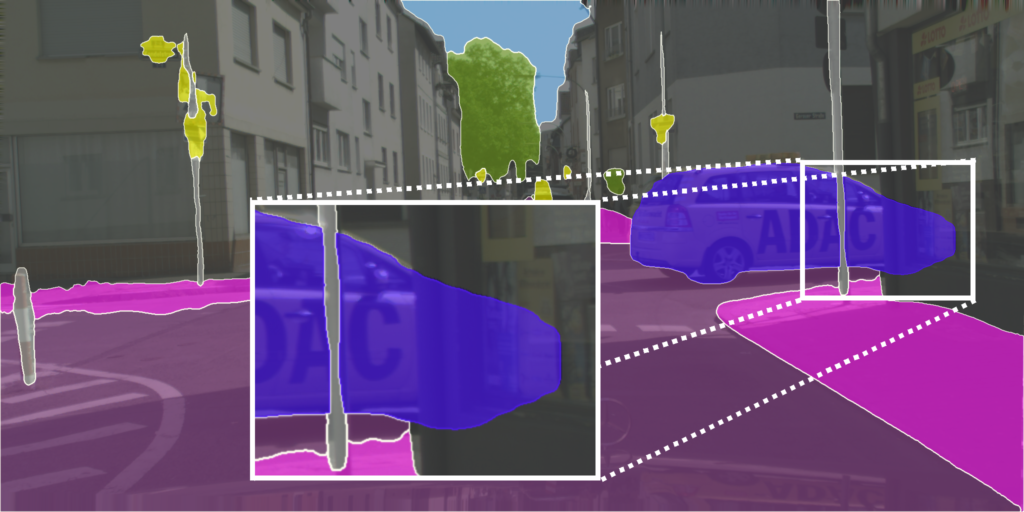}~
\includegraphics[width=0.49\columnwidth]{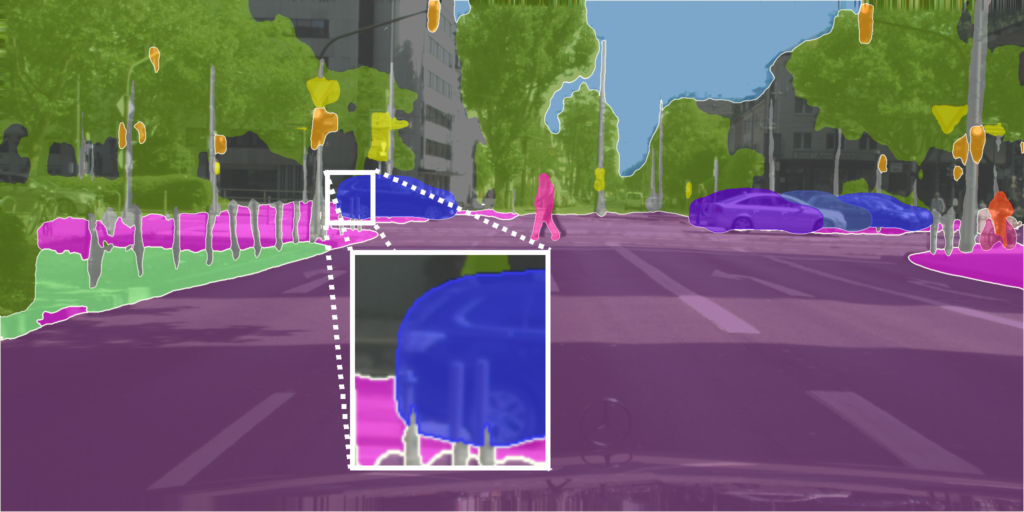}
\vspace{-0.7cm}
\caption{\textbf{Failure cases.} Left: the car is incorrectly predicted to be in front of the building on the right. Right: the car is incorrectly predicted to be in front of a few poles.}
\label{fig:errors}
\vspace{-0.4cm}
\end{figure}

\noindent\textbf{Limitations.} \cref{fig:errors} presents a few sequences where our model mispredicts the relative location of foreground and background components. The noisiness of the input point clouds can introduce error in depth reasoning, especially for far away objects which have similar depth. The fact that we only use one depth value for a foreground instance can  introduce errors for larger objects. %
Similar to IndRNN-Stack, the   method struggles with  instance detection and tracking errors as we assume these inputs to be correct. %

\begin{table}[t]
\setlength{\tabcolsep}{3pt}
    \footnotesize
      \centering
      \begin{tabular}{lcccccc}
    \toprule
    & \multicolumn{3}{c}{$\Delta t = 3$} & \multicolumn{3}{c}{$\Delta t = 9$} \\
    & PQ & SQ & RQ & PQ & SQ & RQ \\
    \midrule
    \textbf{Ours} & $\textbf{50.2}$ & $\textbf{75.7}$ & $\textbf{64.3}$ & $\textbf{37.6}$ & $71.4$ & $\textbf{49.5}$ \\
    1) w/o difference attention & $49.9$ & $75.6$ & $64.0$ & $36.8$ & $\textbf{71.6}$ & $48.3$\\
    2) w/o auxiliary encoder loss & $49.1$ & $75.3$ & $63.0$ & $36.5$ & $71.5$ & $47.9$ \\
    3) w/o refinement & $49.9$ & $75.6$ & $63.9$ & $36.4$ & $71.0$ & $48.0$ \\
    4) w/ ORB-SLAM odometry & $49.6$ & $\textbf{75.7}$ & $63.5$ & $37.2$ & $71.5$ & $49.0$ \\
    \midrule
    w/ ground truth future odometry  & $50.5$ & $75.8$ & $64.7$ & $39.7$ & $72.0$ & $52.1$ \\
    \bottomrule
      \end{tabular}
      \vspace{-0.3cm}
      \caption{\textbf{Validating our design choices} using Cityscapes. Higher is better for all metrics. All approaches use predicted future odometry unless otherwise specified.}
      \label{tab:ablations}
      \vspace{-0.5cm}
 \end{table}

\begin{table*}[t]
    \scriptsize
    \centering
    \begin{tabular}{l ccc ccc ccc c ccc ccc}
         \toprule
         & \multicolumn{3}{c}{All} & \multicolumn{3}{c}{Things} & \multicolumn{3}{c}{Stuff} && \multicolumn{3}{c}{All} & \multicolumn{3}{c}{Things}\\
         & PQID & SQID & RQID & PQID & SQID & RQID & PQID & SQID & RQID & & PQ & SQ & RQ & PQ & SQ & RQ\\
         \midrule
         Pan. Deeplab $\dagger$ & $64.0$ &  $84.9$ &  $74.4$ & $60.9$ &  $79.5$ &  $76.3$ & $64.6$ & $85.9$ & $74.0$ & & $64.0$ & $84.9$ & $74.4$ & $60.9$ &  $79.5$ &  $76.3$ \\
         \midrule
         Pan. Deeplab$*$ & $37.5$ &  $\mathbf{75.8}$ & $47.5$ & $16.1$ & $\mathbf{70.8}$ & $21.5$ & $41.4$ & $76.7$ & $52.2$ & & $37.7$ & $75.5$ & $47.8$ &$17.5$ & $\mathbf{68.8}$ & $23.8$ \\
         Flow & $40.8$ & $75.7$ & $51.6$ & $19.1$ & $69.9$ & $25.7$ & $44.7$ & $76.7$ & $56.3$ & & $40.9$ & $75.4$ & $51.7$ & $19.8$ & $68.5$ & $26.9$ \\
         IndRNN-Stack & $45.1$ & $73.9$ & $57.4$ & $24.1$ & $68.7$ & $32.8$ & $49.0$ & $74.9$ & $61.9$ & & $45.3$ & $73.8$ & $57.7$ & $25.1$ & $68.1$ & $34.5$ \\ 
         \textbf{Ours} & $45.1$ & $73.9$ & $57.2$ & $25.8$ & $69.9$ & $34.3$ &$48.6$ & $74.7$ & $61.4$ & & $45.2$ & $73.8$ & $57.5$ & $26.6$ & $69.0$ & $35.8$ \\
         \midrule
         IndRNN-Stack$\ddagger$ & $48.3$ & $75.1$ & $61.1$ & $24.7$ &  $68.9$ &  $33.3$ & $\mathbf{52.6}$ & $\mathbf{76.2}$ &$\mathbf{66.1}$ & & $48.5$ & $\mathbf{75.0}$ & $61.3$ &$25.8$ &  $68.3$  & $35.1$ \\
         \textbf{Ours}$\ddagger$ & $\mathbf{48.5}$ & $75.1$ & $\mathbf{61.3}$ & $\mathbf{26.2}$ & $69.3$ & $\mathbf{34.9}$ & $\mathbf{52.6}$ & $76.1$ & $\mathbf{66.1}$ & & $\mathbf{48.7}$ &$\mathbf{75.0}$ & $\mathbf{61.5}$ & $\mathbf{26.9}$ &  $\mathbf{68.8}$ &  $\mathbf{36.2}$ \\
         \bottomrule
    \end{tabular}%
    \vspace{-0.25cm}
    \caption{Metrics computed on the AIODrive dataset (0.5 second forecast). $\dagger$: Oracle; $*$: Applied to last seen frame; $\ddagger$: use ground-truth depth inputs. IndRNN-Stack and our models use ground-truth odometry input. 
    }
    \label{tab:aiodrive}
    \vspace{-0.25cm}
\end{table*}

\begin{figure*}[t]
\centering
\footnotesize
\setlength\tabcolsep{0.5pt}
\renewcommand{\arraystretch}{0.5}
\begin{tabular}{cccc}
\rotatebox{90}{\hspace{0.28cm}IndRNN-Stack} & \includegraphics[width=0.32\textwidth]{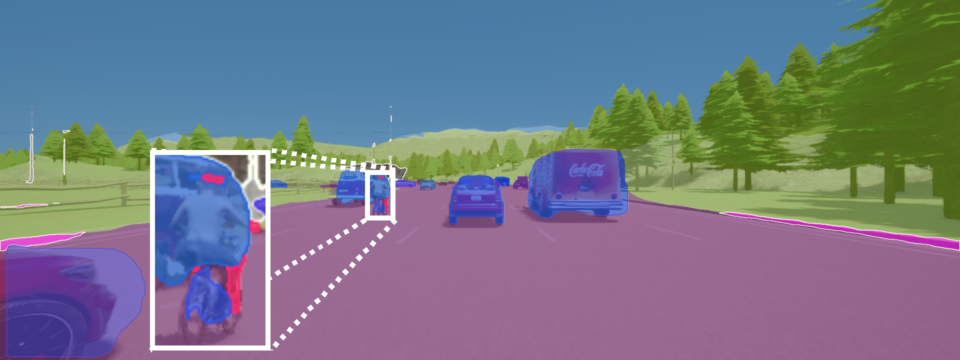}& \includegraphics[width=0.32\textwidth]{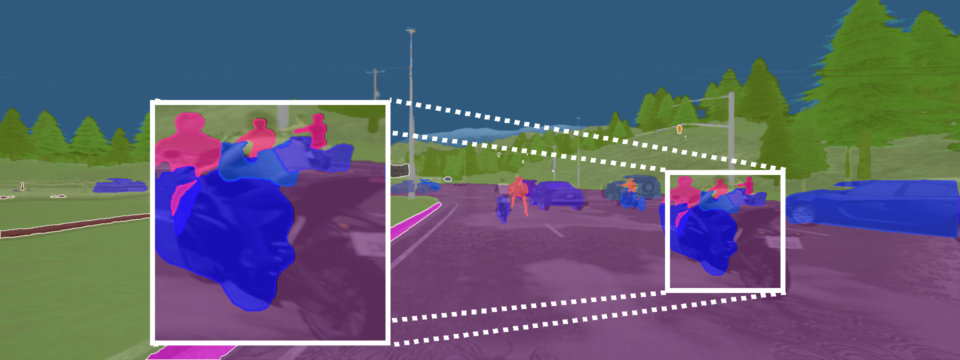} & \includegraphics[width=0.32\textwidth]{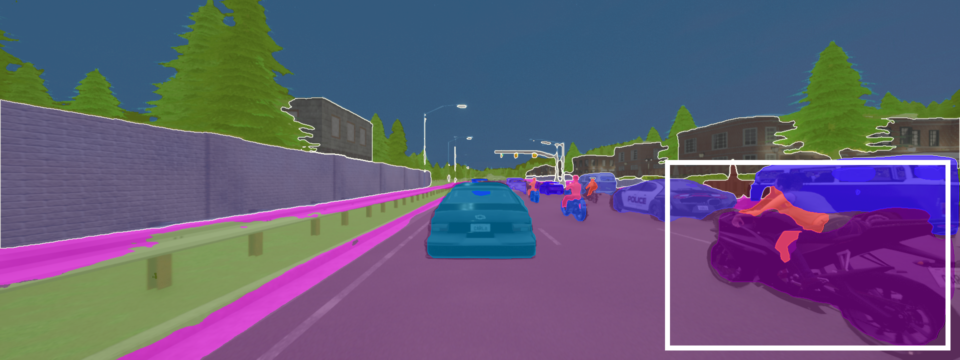} \\ 
\rotatebox{90}{\hspace{0.8cm}\textbf{Ours}} & \includegraphics[width=0.32\textwidth]{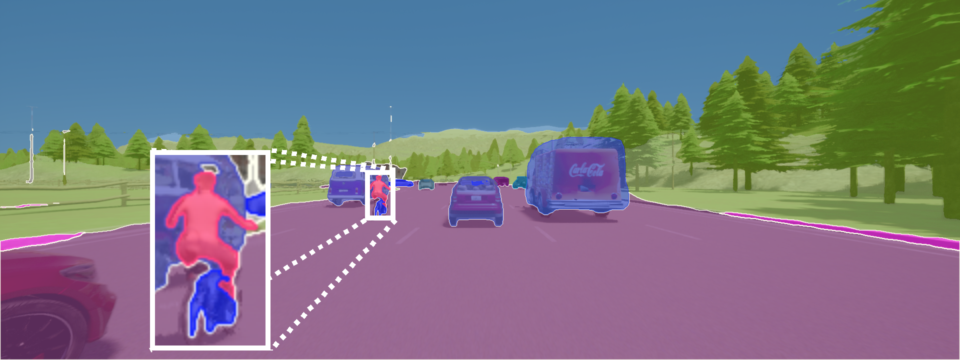}& \includegraphics[width=0.32\textwidth]{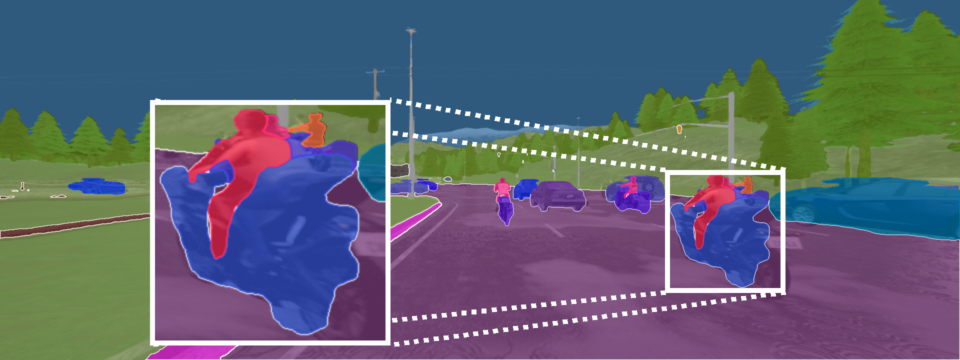}& \includegraphics[width=0.32\textwidth]{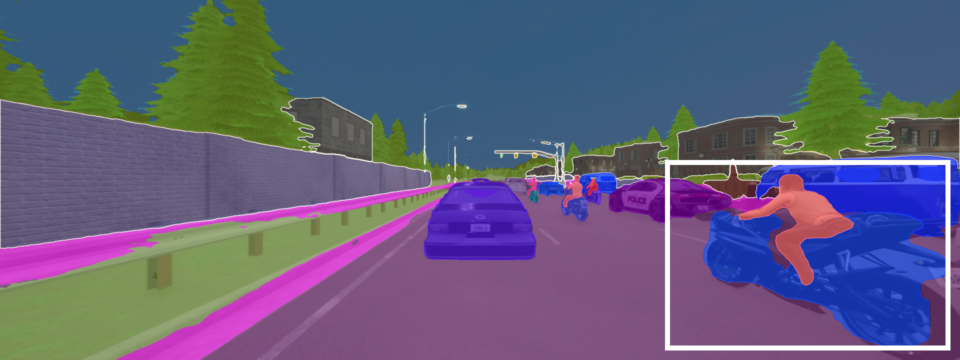}
\end{tabular}
\vspace{-0.25cm}
\caption{Panoptic segmentation forecasts on the AIODrive dataset. Our approach better captures the relationships between  cyclists/their vehicles than IndRNN-Stack, which cannot model these relationships. Also, our model produces higher-fidelity pedestrian instance masks.}
\label{fig:aiodrive_panoptic_viz}
\vspace{-0.5cm}
\end{figure*}

\noindent\textbf{Ablations.}
\cref{tab:ablations} summarizes  results studying the impact of  modeling decisions. 1)~\emph{w/o difference attention} uses standard dot product attention for all transformers in place of the difference attention module developed in \cref{transformer}. Our full model's superior performance over 1) demonstrates that the difference attention model is able to better reason about instance motion. 2)~\emph{w/o auxiliary encoder loss} trains the forecasting model without applying a loss to the velocity output from \cref{eq:enc_vel}. This leads to worse results, and shows that the auxiliary loss helps bias the encoder representations to encode motion information  useful for forecasting. 3)~\emph{w/o refinement} does not use the refinement head, and instead stacks  foreground predictions on top of  background predictions, following Graber \etal~\cite{graber2021panoptic}. This leads to  missed background objects which are occluded by foreground predictions, hence a drop in results. 4)~\emph{w/ ORB-SLAM odometry} uses input odometry obtained from \cite{ORBSLAM3_2020}, and shows that our method also works with odometry obtained from image data. The final ablation demonstrates that  access to more accurate future camera motion leads to  improvements.

 \vspace{-0.2cm}
\subsection{AIODrive}
\label{sec:aiodrive}
\vspace{-0.2cm}
\noindent\textbf{Data.} The AIODrive dataset~\cite{Weng2020_AIODrive} contains a large number of synthetically generated traffic scenarios and provides many inputs and annotations, including stereo images, LiDAR, ground-truth depth maps, panoptic segmentations, and more. The use of a simulator to obtain data and annotations results in AIODrive containing panoptic segmentation annotations, including instance tracks, for all frames. Here, we use the subset of the labels corresponding to Cityscapes classes, consisting of 2 ``things'' and 11 ``stuff'' classes. We use 5 frames of input and forecast the 5th frame into the future (corresponding to a 0.5s forecast). Additional details can be found in \cref{app:aiodrive}.

\noindent\textbf{Metrics.} In addition to previously used metrics, we introduce metrics which account for object identity. Specifically, we evaluate using PQID, SQID, and RQID, which require matches computed between predicted and ground-truth objects to have the same instance ID. These metrics are more appropriate for the forecasting setting due to the fact that the previously used metrics can mark matches between different instances as true positives, meaning the motion of an instance was incorrectly predicted but the metric did not properly evaluate this. Note that we cannot compute these metrics on Cityscapes, as that data only contains annotations for a single frame per sequence. 

\noindent\textbf{Results.} The results for all models on the panoptic segmentation forecasting task on AIODrive are presented in \cref{tab:aiodrive}. Because the All PQID is averaged over 2 ``things'' classes and 11 ``stuff'' classes, this metric is biased towards ``stuff'' performance. Hence, All PQID is comparable between IndRNN-Stack and our model. However, the differences are much clearer on the ``things'' metrics, as our approach is better able to reason about the motion of individual object instances. Furthermore, there is a small drop in performance between PQID and PQ, indicating that some of the true positive matches found when computing PQ are between incorrect ground-truth instances. 
\cref{fig:aiodrive_panoptic_viz} shows results for our method and IndRNN-Stack on AIODrive. Our approach produces better forecasts for cyclists and their bikes, due to the use of difference attention. %

\vspace{-0.2cm}
\section{Conclusion}
\vspace{-0.2cm}
We introduce a new model for  panoptic segmentation forecasting. It uses difference attention  which we find to be  more suitable to forecasting than standard attention as it can reason about velocities and acceleration. A new refinement head also merges predictions based on depth. This improves  prior work on all panoptic forecasting metrics.

\noindent\textbf{Acknowledgements:} This work is supported in part by NSF \#1718221, 2008387, 2045586, 2106825, MRI \#1725729, NIFA 2020-67021-32799 and Cisco Systems Inc.\ (CG 1377144 - thanks for access to Arcetri).

{
    \small
    \bibliographystyle{ieee_fullname}
    \bibliography{macros,main}
}

\appendix
\onecolumn
\setcounter{page}{1}

\section*{\Large\centering Supplementary Material: \\Joint Forecasting of Panoptic Segmentations with Difference Attention}

This appendix is structured as follows: \cref{app:bgmodel} details the background prediction approach which we use to obtain preliminary background class predictions. \cref{app:locappfeat} provides specific model architectural details for the forecasting transformer encoder and decoder. \cref{app:agentawareattention} explains in detail the agent-aware attention approach we use which allows for identity information to be encoded in the model. \cref{app:foregroundforecastinglosses} describes the specific losses computed during training of the foreground forecasting model. \cref{app:depthcompletion} presents the model architecture used by the depth completion model introduced in \cref{refine}. \cref{app:implementation} describes additional details of implementation and model training.
\cref{app:aiodrive} contains additional information about the AIODrive dataset and experiments. \cref{app:instseg} contains instance segmentation forecasting experimental results for Cityscapes. \cref{app:semseg} contains semantic segmentation forecasting experimental results for Cityscapes. \cref{app:additionalviz} presents additional model visualizations on Cityscapes for both the short- and mid-term settings. \cref{app:full_metrics} contains the full per-class breakdown of the panoptic segmentation metrics presented in \cref{tab:panoptic}. \cref{app:code} describes the major code libraries used to implement our model. Finally, \cref{app:impact} discusses potential negative societal impacts that could arise from the implementation of this work in practice.

\section{Background Model}
\label{app:bgmodel}
In this work, we utilize the background semantic prediction model introduced by Graber \etal~\cite{graber2021panoptic}. This approach lifts background semantics into a 3D point cloud using the estimated input depth, transforms the point cloud based on camera movement, projects to the image plane, and refines the projected semantics using a semantic segmentation model. Formally, this model estimates the semantics of background object classes for unseen future frame $T+F$ as 
\begin{equation}
    \widehat{m}_{T+F}^B = \text{BGRef}(\{\text{proj}(m_t,d_t,K,H_t,u_t)\}_{1:T}),
\end{equation}
where $K$ represents camera intrinsic parameters, $H_t$ is the $6$-dof camera transform from input frame $t$ to target frame $T+F$, $m_t$ is the semantic segmentation for frame $t$ which is obtained from a pre-trained model, $d_t$ is the input depth map at time $t$, and $u_t$ denotes the coordinates of all of the pixels in $m_t$ which correspond to background semantic classes. \emph{Proj} refers to the step which creates the sparse reprojected semantic map for frame $T+F$ given inputs for frame $t$, and \emph{BGRef} refers to the background refinement model which produces a complete background prediction from the output of \emph{Proj}.

The first step of the background model is to produce reprojected semantic point clouds $(\tilde{m}_t^B,\tilde{d}_t^B)$ which are processed by \emph{BGRef}. These are obtained for each time $t \in \{1, \dots, T\}$ by applying \emph{Proj} to the corresponding input frame $I_t$. Given per-pixel semantic prediction $m_t$ and depth map $d_t$, \emph{Proj} back-projects, transforms, and reprojects the pixels from input frame $t$ to target frame $T+F$. This process is summarized as
\begin{align}
    \begin{bmatrix} x_t \\ y_t \\ z_t \end{bmatrix} = H_t \begin{bmatrix} K^{-1} \begin{bmatrix} u_t \\ 1 \end{bmatrix} \text{diag}(d_t) \\ 1 \end{bmatrix}, \\
    \begin{bmatrix} u_{T+F} \\ 1 \end{bmatrix} = K \begin{bmatrix}  x_t / z_t \\ y_t / z_t \\ 1 \end{bmatrix}, \\
    \tilde{m}_t^B(u_{T+F}) = m^B_t(u_t), \\
    \tilde{d}_t^B(u_{T+F}) = z_t,
\end{align}
where $u_t$ is a vector whose entries dictate the pixel locations in $m_t$ which correspond to background object classes and $u_{T+F}$ is the vector which contains the location of these pixels in the target frame at time $T+F$. During this, we maintain the semantic class obtained from $m_t$ and the projected depth of each pixel location. Whenever multiple pixels $u_t$ from an input frame are projected to the same pixel $u_{T+F}$ in the target frame, the depth and the semantic label of the pixel with the smallest depth is kept, as it is closest to the camera.

Given reprojected semantics $\tilde{m}_t^B$ and depths $\tilde{d}_t^B$ from the previous step, the background refinement model is tasked with predicting a final semantic output. This is done by concatenating the input from all frames and feeding them into a semantic segmentation model, which can be described as
\begin{align*}
    \widehat{m}^{\text{Prob}}_{T+F} &= \text{BGRef}([\{\tilde{m}_t^B(u_{T+F}) \tilde{d}_t^B(u_{T+F})\}_{1:T}])\\
    \widehat{m}_{T+F}^B &= \argmax_c(\widehat{m}^{\text{Prob}}_{T+F}),
\end{align*}
where $\widehat{m}^{\text{Prob}}_{T+F}\in \Delta_{C_\text{BG}}^{H \times W}$ represents the $C_\text{BG}$-dimensional output probability map per pixel, one for each background class, and the final output $\widehat{m}_{T+F}^B$ is obtained per-pixel by choosing the class with the largest probability.

The refinement network is trained using the cross-entropy loss
\begin{align}
    \mathcal{L}_\text{bf} \coloneqq \tfrac{1}{\sum_{x,y}\mathbf{1}^{\text{bg}}_{T+F}[x,y]} \sum_{x,y} \mathbf{1}^\text{bg}_{T+F}[x,y] \sum_{c} m^{B*}_{T+F}(x,y,c) \log\left(\widehat{m}^{\text{Prob}}_{T+F}(x,y)\right). 
\end{align}
Here, $\mathbf{1}^{\text{bg}}_{T+F}[x,y]$ is an indicator function specifying whether pixel coordinates $(x,y)$ correspond to background semantic classes for frame $T+F$, and $m^{i*}_{T+F}(x,y,c)=1$ if the correct class for pixel $(x,y)$ is $c$ and $0$ otherwise. 
For all experiments presented in this work, we use the specific background prediction model trained by Graber \etal~\cite{graber2021panoptic}. Further implementation details related to model architecture and training can be found in the Appendix of \cite{graber2021panoptic}.

\section{Architecture details for Forecasting Transformer Encoder and Decoder}
\label{app:locappfeat}
The feature model $f_\text{Loc}$ processes input locations $\mathbf{x}_t^i$, appearances $\mathbf{r}_t^i$, instance classes $c^i$, odometry $o_t$, and time $t$ to produce an embedding $\bar{\mathbf{x}}^i_{\text{Loc},t}$ which is processed by the transformer FTE. $f_\text{Loc}$ can be fully specified by the following model components:
\begin{align}
    {\mathbf{x}_t'}^{i} &= f_b([\mathbf{x}_t^i, \text{onehot}(c^i)]), \\
    {\mathbf{r}_t'}^i &= \text{AvgPool}(f_f(\mathbf{r}_t^i)), \\
    \bar{\mathbf{x}}^i_{\text{Loc},t} &= f_\text{e2}([f_\text{e1}([{\mathbf{x}_t'}^i, {\mathbf{r}_t'}^i, o_t]), \tau_t]).
\end{align}
First, an initial location embedding ${\mathbf{x}_t'}^{i}$ is produced, where $f_b$ is a linear layer and \emph{onehot} represents a vector whose $c^i$-th element is set to one and whose other entries are set to zero. Similarly, initial appearance embedding ${\mathbf{r}_t'}^i$ is produced, where $f_f$ is a small convolutional network and \emph{AvgPool} averages the result over the spatial dimensions. These two embeddings are concatenated with odometry $o_t$, passed through linear layer $f_{e1}$, concatenated with temporal encoding $\tau_t$, and passed through the final linear layer $f_{e2}$. Specifically, the temporal encoding $\tau^t \in \mathbb{R}^{d_\tau}$ provides information to the model about the temporal location of the given instance in the sequence and whose $k$-th element is defined as
\begin{equation}
    \tau^t(k) = \begin{cases}
    \sin(t/1000^{k/d_\tau}), & k \text{ is even}\\
    \cos(t/1000^{(k-1)/d_\tau}), & k \text{ is odd}\\
    \end{cases}.
\end{equation}
Here $d_\tau$ is the size of the temporal encoding and is set to $256$ everywhere in this work. All linear layers in $f_\text{Loc}$ have an output embedding size of $256$, and $f_f$ contains two 2D convolutional layers with a kernel size of $3$, output channel size of $256$, and ReLU activations after each.

The feature model $f_\text{App}$ produces appearance embedding $\bar{\mathbf{x}}^i_{\text{App},t}$ as a function of the input appearances $\mathbf{r}_t^i$ as well as input time $t$, and can be fully specified by the following model components:
\begin{equation}
    \bar{\mathbf{x}}_{\text{App},t}^i = f_\text{ae2}([f_\text{ae1}(\mathbf{r}_t^i), \tilde{\tau}_t]),
\end{equation}
where $f_\text{ae1}$ is a $3 \times 3$ convolutional layer with output dimension $256$, $f_\text{ae2}$ is a $1 \times 1$ convolutional layer with output dimension $256$, and $\tilde{\tau}_t \in \mathbb{R}^{d_\tau \times 14 \times 14}$ is equivalent to $\tau_t$ copied across spatial dimensions to match the size of $\mathbf{r}_t^i$. 

The location transformer encoder $\text{FTE}_\text{Loc}$ consists of two stacks of transformer encoder modules as originally defined in \cite{vaswani2017attention} consisting of layer norm, multi-head self-attention, feed-forward networks, and residual connections. Specifically, all transformers in this work use the Pre-LN construction \cite{xiong2020layer}, where the Layer Norm module is placed before the multi-head attention and feed-forward network, as we observed improved convergence. As specified in \cref{foreground}, the multi-head attention modules use both difference attention (\cref{transformer}) and agent-aware attention (\cref{app:agentawareattention}). The embedding dimension of all keys, queries, and values as well as the output $\mathbf{h}_{\text{Loc},t}^i$ is $256$, the hidden dimension of feedforward modules is $512$, the dropout rate used is $0.1$, and the number of heads used for multi-head attention is $8$.

The appearance transformer encoder $\text{FTE}_\text{App}$ additionally consists of two stacks of transformer encoder modules. However, unlike $\text{FTE}_\text{Loc}$, the standard dot-product attention formulation is used, and all linear projections in both the multihead attention modules as well as the feedforward network are replaced with 2D convolutional layers with a filter size of $3 \times 3$. All embeddings maintain the same spatial dimensions of $14 \times 14$ during computation, the channel dimension used is $256$, the hidden channel dimension of the feedforward modules is $512$, the dropout rate used is $0.1$, and the number of heads used for multi-headed attention is $8$.

Note, for readability we formulate all models assuming every instance $i$ is present at every input time step $t \in \{1, \dots, T\}$. However, in practice, some instances will not be present in some input time steps due to occlusions or instances entering/leaving the frame, \ie, there are instances $i$ and input frames $t$ for which $p_t^i = 0$. For all $i, t$ such that $p_t^i = 0$, we do not compute $\bar{\mathbf{x}}_{\text{Loc},t}^i$ or  $\bar{\mathbf{x}}_{\text{App},t}^i$ since there are no inputs from which we can compute these. Consequentially, neither $\text{FTE}_\text{Loc}$ nor $\text{FTE}_\text{App}$ receive input representing instance $i$ for time $t$ and thus do not produce encoder representations $\textbf{h}^i_{\text{Loc},t}$ and $\textbf{h}^i_{\text{App},t}$ for them.

The decoder location feature model $\tilde{f}_\text{Loc}$ produces the feature representation $\tilde{\mathbf{x}}_{\text{Loc},t}^i$ containing information about the most recently predicted location, odometry, and the corresponding instance class.  $\tilde{f}_\text{Loc}$ can be fully specified by the following model components:
\begin{align}
    {\textbf{x}''}_t^{i} &= f_{\text{d1}}([\widehat{\textbf{x}}_{t-1}^i, \text{onehot}(c^i), o^t), \\
    \tilde{\mathbf{x}}_{\text{Loc},t}^i &= f_\text{d2}([{\textbf{x}''}_t^{i}, \tau_{t}]).
\end{align}
First, an initial representation ${\textbf{x}''}_t^{i}$ is computed from the previous location prediction $\widehat{\textbf{x}}_{t-1}^i$ using linear layer $f_\text{d1}$, corresponding instance class $c^i$, and odometry $o_t$. This is concatenated with temporal encoding $\tau_t$ and passed through a second linear layer $f_\text{d2}$. Both $f_\text{d1}$ and $f_\text{d2}$ use output dimension equal to $256$.

The decoder appearance feature model $\tilde{f}_\text{App}$ produces the feature representation $\tilde{\mathbf{x}}_{\text{App},t}^i$ containing information about the most recently predicted appearance. $\tilde{f}_\text{App}$ can be fully specified by the following model components: 
\begin{equation}
    \tilde{\mathbf{x}}_{\text{App},t}^t = f_\text{ad2}([f_\text{ad1}(\widehat{\textbf{r}}_{t-1}^i), \tilde{\tau}_t]),
\end{equation}
where $f_\text{ad1}$ and $f_\text{ad2}$ are convolutional layers with the same structure as $f_\text{ae1}$ and $f_\text{ae2}$, respectively.

Both the location and appearance transformer decoders $\text{FDE}_\text{Loc}$ and $\text{FDE}_\text{App}$ use the same construction and hyperparameters as their encoder counterparts. The primary difference is that they are transformer decoders as defined in \cite{vaswani2017attention} and hence additionally introduce cross attention layers which operate on the encoder representations $\{\mathbf{h}_{\text{Loc},t}^i\}_{1:T}^{1:N}$ and $\{\mathbf{h}_{\text{App},t}^i\}_{1:T}^{1:N}$, respectively. Output decoder representations $\tilde{\mathbf{h}}_{\text{Loc},t}^i$ and $\tilde{\mathbf{h}}_{\text{Loc},t}^i$ are computed autoregressively; \eg, previous predictions $\{\widehat{\textbf{x}}^i_t\}_{T:t'-1}^{1:N}$ for times $T$ through $t'-1$ are used to compute the outputs $\{ \tilde{\textbf{h}}_{\text{Loc},t'}^i\}$. These embeddings are then used to produce $\{\widehat{\textbf{x}}_{t'}^i\}^{1:N}$ for time $t'$, and these new predictions are fed back into the model to produce output for the next time step $t'+1$, and so on. Decoder attention is masked to maintain causality, \ie, embeddings representing a given time $t$ are prevented from attending to representations for future time steps $t' > t$.

$f_\text{LocOut}$, $f_\text{POut}$, and $f_\text{vel}$ are all $3$-layer multilayer perceptrons with hidden sizes $[512, 256]$ and ReLU activations. $f_\text{AppOut}$ is a $3 \times 3$ convolutional layer.

\section{Agent-aware Attention}
\label{app:agentawareattention}
Due to their permutation-invarance with respect to their inputs, transformers do not have the inherent capacity to reason about the identity of the entities corresponding to input trajectories. To address this problem, Yuan \etal~\cite{yuan2021agentformer} introduced agent-aware attention. This approach allows transformers to encode the identity of its inputs within the model, which makes it easier  for these models to reason about the trajectories of individual entities and leads to better forecasting performance.

Let $\mathbf{X}_\text{self} \in \mathbb{R}^{M_1 \times d}$ and $\mathbf{X}_\text{other} \in \mathbb{R}^{M_2 \times d}$ of lengths $M_1$ and $M_2$, respectively, be the input sequences with embedding dimension $d$. For self-attention, both input sequences are the same and represent the input trajectories of a number of agents, while for cross-attention, the first input sequence corresponds to a future trajectory forecast and the second corresponds to input trajectories. The agent-aware attention output $\mathbf{Y} \in \mathbb{R}^{M_1 \times d}$ is then computed as
\begin{align}
    \mathbf{Z} &= \mathbf{M} \odot (\mathbf{Q}_\text{agent} \mathbf{K}^T_\text{agent}) + (1 - \mathbf{M})\odot(\mathbf{Q}_\text{context}\mathbf{K}^T_\text{context}),\\ 
    \mathbf{Y} &= \text{softmax}\left(\mathbf{Z} / \sqrt{d}\right) \mathbf{V},
\end{align}
where $\odot$ represents element-wise multiplication. Specifically, agent-aware attention first computes two sets of keys $\mathbf{K}_\text{agent} = f_{K,\text{agent}}(\mathbf{X}_\text{other})$, $\mathbf{K}_\text{context} = f_{K,\text{context}}(\mathbf{X}_\text{other})$ and queries $\mathbf{Q}_\text{agent} = f_{Q,\text{agent}}(\mathbf{X}_\text{self})$, $\mathbf{Q}_\text{context} = f_{Q,\text{context}}(\mathbf{X}_\text{self})$ from the original inputs. It then computes two sets of attention scores from the \emph{agent} keys/queries and from the \emph{context} keys/queries and selects between them using mask $\mathbf{M} \in \{0, 1\}^{M_1 \times M_2}$. This mask encodes identity information: $\mathbf{M}_{ij} = 1$ if entity $i$ in the first input sequence and entity $j$ in the second input sequence correspond to the same agent, and $\mathbf{M}_{ij} = 0$ otherwise. In other words, two sets of attention parameters are computed, and one set is used for input pairs corresponding to the same agent while the other is used for all pairs corresponding to different agents, \ie, the context for this agent. Value aggregation proceeds as in standard attention from this step.

We additionally use agent-aware attention within the difference attention module defined in \cref{transformer}. This is implemented in a similar fashion, where separate attention parameters are computed for input pairs corresponding to the same agent and for input pairs corresponding to different agents. We formally specify this as
\begin{align}
    \mathbf{Z} &= \mathbf{M} \odot \left(\mathbf{Q}_\text{agent} \mathbf{K}^T_{R,\text{agent}} - \mathbf{1}_{M_1 \times 1} \text{diag}\left(\mathbf{K}_{B,\text{agent}}\mathbf{K}_{R,\text{agent}}^T\right)^T\right) + \\
    & \hspace{0.5cm}(1 - \mathbf{M})\odot\left(\mathbf{Q}_\text{context} \mathbf{K}^T_{R,\text{context}} - \mathbf{1}_{M \times 1} \text{diag}\left(\mathbf{K}_{B,\text{context}}\mathbf{K}_{R,\text{context}}^T\right)^T\right),\\ 
    \mathbf{Y} &= \text{softmax}\left(\mathbf{Z} / \sqrt{d}\right) \mathbf{V}_O - \mathbf{V}_S,
\end{align}
with $\mathbf{K}_{R,\text{agent}} = f_{K,R,\text{agent}}(\mathbf{X}_\text{other})$, $\mathbf{K}_{R,\text{context}} = f_{K,R,\text{context}}(\mathbf{X}_\text{other})$, $\mathbf{K}_{B,\text{agent}} = f_{K,B,\text{agent}}(\mathbf{X}_\text{other})$, $\mathbf{K}_{B,\text{context}} = f_{K,B,\text{context}}(\mathbf{X}_\text{other})$, $\mathbf{V}_O = f_{V_O}(\mathbf{X}_\text{other})$, and $\mathbf{V}_S = f_{V_S}(\mathbf{X}_\text{self})$.

\section{Losses for Foreground Forecasting}
\label{app:foregroundforecastinglosses}

The loss used by the foreground forecasting model are
\begin{equation}
    \mathcal{L}_\text{FG} = \mathcal{L}_\text{Loc} + \mathcal{L}_\text{P} + \mathcal{L}_\text{App} + \mathcal{L}_\text{Vel}.
\end{equation}
The location loss $\mathcal{L}_\text{Loc}$ trains the bounding box predictions $\widehat{\textbf{x}}_\text{Box,t}^i \coloneqq [\widehat{x}_{0,t}, \widehat{y}_{0,t}, \widehat{x}_{1,t},  \widehat{y}_{1,t}]$ and depth predictions $\widehat{d}_t^i$ to match the target boxes $\textbf{x}_\text{Box,t}^{*i}$ and depths $d_t^*$. This is specified as
\begin{equation}
    \mathcal{L}_\text{Loc} \coloneqq \frac{1}{\sum_{i=1}^N\sum_{t=T+1}^{T+F} p^{*i}_t } \sum_{i=1}^N\sum_{t=T+1}^{T+F}p_t^{*i}\left(\lambda_1\text{SmoothL1}(\widehat{\textbf{x}}_{\text{Box},t}^i, \textbf{x}_{\text{Box},t}^{*i}) + \lambda_2\text{SmoothL1}(\widehat{d}_t^i, d^{i*}_t) + \lambda_3\text{IoU}(\widehat{\textbf{x}}_{\text{Box},t}^i, \textbf{x}_{\text{Box},t}^{*i})\right),
\end{equation}
where $p_t^{*i}$ is ground-truth presence, \ie, equals $1$ if instance $i$ is present in frame $t$ and $0$ otherwise, IoU is bounding box intersection-over-union, SmoothL1 is the function
\begin{align}
    \text{SmoothL1}(\mathbf{a}, \mathbf{b}) &\coloneqq \sum_j \text{SmoothL1Fn}(\textbf{a}_j, \textbf{b}_j), \\
    \text{SmoothL1Fn}(a,b) &\coloneqq \begin{cases}
        \tfrac{1}{2}(a-b)^2, &\text{if }|a-b| < 1,\\
        |a - b| - \tfrac{1}{2} &\text{otherwise}
    \end{cases},
\end{align}
and coefficients $\lambda_1=1$, $\lambda_2=10$, $\lambda_3=100$ are used to balance the magnitudes of the losses.

The presence loss $\mathcal{L}_\text{P}$ trains the presence predictions $\widehat{p}_t^i \in \mathbb{R}$ to correctly indicate whether a given instance $i$ is present in frame $t$, and is computed as
\begin{equation}
    \mathcal{L}_\text{P} \coloneqq \frac{\lambda_4}{NF} \sum_{i=1}^N\sum_{t=T+1}^{T+F}p_t^{*i}\log \sigma (\widehat{p}_t^i) + (1 - p_t^{*i})\log(1 - \sigma(\widehat{p}_t^i)),
\end{equation}
where $\sigma$ is the sigmoid function and $\lambda_4=10$.

The appearance loss $\mathcal{L}_\text{App}$ trains the appearance predictions $\widehat{\textbf{r}}_t^i$ for instance $i$ at frame $t$ to match the target features $\textbf{r}_t^{*i}$ extracted for this instance at frame $t$, and consists of the mean-squared error of the features for all valid instance/time pairs, \ie,
\begin{equation}
    \mathcal{L}_\text{App} \coloneqq \frac{\lambda_5}{\sum_{i=1}^N\sum_{t=T+1}^{T+F} J p^{*i}_t } \sum_{i=1}^N\sum_{t=T+1}^{T+F}\sum_{j=1}^J p_t^{*i}(\widehat{\textbf{r}}_{j,t}^i- \textbf{r}_{j,t}^{*i})^2,
\end{equation}
where $j$ indexes over all spatial dimensions of the feature tensors, $J = 256 \times 14 \times 14$ is the total number of elements of the feature tensors, and $\lambda_5=10$ .

The encoder velocity loss $\mathcal{L}_\text{Vel}$ trains the velocity predictions $\widehat{\textbf{v}}_{\text{E},t}^i \in \mathbb{R}^4$ to match the ground-truth velocities $\textbf{v}_{t}^{*i} \coloneqq \textbf{x}_{t+1}^{*i} - \textbf{x}_{t}^{*i}$, and is computed as 
\begin{equation}
    \mathcal{L}_\text{Vel} \coloneqq  \frac{\lambda_6}{\sum_{i=1}^N\sum_{t=1}^{T} p^{*i}_t p^{*i}_{t+1} } \sum_{i=1}^N\sum_{t=1}^{T}p_t^{*i}p_{t+1}^{*i}\text{SmoothL1}(\widehat{\textbf{v}}_{\text{E},t}^i, \textbf{v}_{t}^{*i}),
\end{equation}
where $\lambda_6=1$.

\section{Depth Completion Model}
\label{app:depthcompletion}
The depth completion model operates on noisy and incomplete reprojected background depth $\tilde{d}^B$  along with depth mask $Q$ and  background class probabilities $\widehat{m}^\text{Prob}$ and produces depth maps $\widehat{d}_\text{Fill}^\text{B}$ and $\widehat{d}_\text{Bias}^B$. This model can be formally represented using the following components:
\begin{align}
    d_1 &= f_\text{dc1}([\tilde{d}^B, Q, \widehat{m}^\text{Prob}]), \\
    d_2 &= d_1 + \text{Upsample}(f_{dc2}(\text{Downsample}(d_1)), \\
    \widehat{d}^B_\text{Fill} &= f_\text{fill}(d_2), \\
    \widehat{d}^B_\text{Bias} &= f_\text{Bias}(d_2).
\end{align}
First, reprojected background depth $\tilde{d}^B$, depth mask $Q$, and predicted background class probabilities $\widehat{m}^\text{Prob}$ are concatenated together and processed with convolutional layer $f_\text{dc1}$ which uses a kernel size of $3$ and has output channel dimension $32$. The output of this, $d_\text{1}$, is downsampled by a factor of $2$ using bilinear interpolation, fed into convolutional network $f_\text{dc2}$, upsampled to the original resolution using bilinear interpolation, and added with $d_\text{1}$ to produce the  second intermediate output $d_\text{2}$. $f_\text{dc2}$ contains $2$ convolutional layers with a kernel size of $3$, output channel dimension of $32$, and a ReLU activation between them. The outputs $\widehat{d}_\text{Fill}^\text{B}$ and $\widehat{d}_\text{Bias}^B$ are then obtained from $d_2$ using convolutional networks $f_\text{fill}$ and $f_\text{bias}$, respectively. Both of these networks contain two convolutional layers with a ReLU activation between them, where the first layer uses a kernel size of $3$ and an output channel size of $32$ and the second layer uses a kernel size of $1$.
The final background depth estimate $\widehat{d}^B$ is obtained from outputs $\widehat{d}_\text{Fill}^\text{B}$ and $\widehat{d}_\text{Bias}^B$ as specified in \cref{eq:bgd}.

\section{Additional Implementation Details}
\label{app:implementation}
The overall approach is trained in two stages: first, the foreground prediction model is trained; afterwards, the corresponding parameters are frozen, and the refinement model is trained. 

The foreground model is trained for $48000$ steps using the ADAM optimizer; the initial learning rate is set to $10^{-4}$, and it is lowered to $10^{-5}$ after $36000$ optimization steps. All odometries $o_t$ are normalized by subtracting the training data set mean and then dividing by training data set standard deviation before being used as input. All location inputs $\mathbf{x}_t^i$ are normalized to lie within $[-1, 1]$; furthermore, location outputs $\widehat{\mathbf{x}}_t^i$ are made at this normalized scale and unnormalized before being used at later stages. During training and inference, forecasts are only predicted for instances present in the most recent input frame, \ie, for instances $i$ such that $p_{T}^i = 1$. During training of the foreground model, ground-truth future odometry is used. During evaluation, unless otherwise noted, the egomotion estimation module described by Graber \etal~\cite{graber2021panoptic} was used to obtain future odometry which was used as input. We use the same odometry representation as Graber \etal~\cite{graber2021panoptic} consisting of a five-dimensional vector containing speed and yaw rate of the ego-vehicle at time $t$ as well its top-down displacement and angular displacement between steps $t$ and $t-1$.

Ablation 4 in this work uses odometry during inference that was obtained using ORB-SLAM3~\cite{ORBSLAM3_2020}. This was run using stereo images, where each sequence of $30$ frames was treated as its own SLAM session providing 6-dof poses for all frames in the sequence.

The refinement model is trained for $24000$ steps using the ADAM optimizer; the initial learning rate is set to $10^{-4}$, and it is lowered to $10^{-5}$ after $18000$ optimization steps. During training, the inputs are scaled to a spatial resolution of $\tfrac{H}{4} \times \tfrac{W}{4}$, and the loss is additionally computed at this scale. During inference, inputs are scaled to the final spatial resolution, \ie, $H \times W$.

To process a Cityscapes sequence, the model needs 560 ms on average using an NVIDIA A6000, which is on par with the 700 ms required by Graber \etal~\cite{graber2021panoptic}. This can be significantly reduced by further engineering effort.

\section{Additional AIODrive Details}
\label{app:aiodrive}
The AIODrive sequences are annotated using 23 object classes. To facilitate comparison against results on the Cityscapes dataset, we only train and evaluate using background classes which are also present in Cityscapes. This leaves $11$ background ``stuff'' classes and $2$ foreground ``things'' classes (the only annotated ``things'' instances in AIODrive are ``vehicles'' and ``pedestrians''). As annotations are only provided for the trainval dataset, we split this into a training dataset containing all annotated sequences for towns 1 through 5 and a validation dataset containing all annotated sequences for town 6. We use $5$ frames as input and forecast the $5$th frame into the future, corresponding to $0.5$ seconds of input and a $0.5$ second forecast (which is comparable to the Cityscapes mid-term setting). 

During both training and evaluation, we only consider instances whose masks have an area of at least 400 pixels in an attempt to filter out distant, imperceptible instances. For evaluation, we use non-overlapping sequences of $10$ frames from each validation sequence. Additionally, the data contains some periods of time with little to no motion, which skews the evaluation metrics artificially high. To ensure that the metrics properly capture the ability of the models to anticipate motion, we filter out validation sequences where the recording vehicle is moving less than 1 m/s at all points in the input sequence and where at least half of the instance mask centers move less than 10 pixels. This leaves $814$ sequences with motion for evaluation purposes. To ensure that the tracking-based metrics can be computed, we use ground-truth instance bounding boxes and ids as input to the forecasting models. 

The base semantic and instance segmentation models are the same as that used for Cityscapes, \ie, MaskRCNN \cite{he2017mask} for instance segmentation and Panoptic Deeplab \cite{Cheng2020panoptic-deeplab} for semantic segmentation. For both, we initialize from the Cityscapes pre-trained model and finetune on AIODrive. For the models that use predicted depth, we use Cascade-Stereo\cite{gu2020cascade} on the stereo input images. We do not finetune the depth model on this dataset.

\section{Cityscapes Instance Segmentation}
\label{app:instseg}
We also evaluate our Cityscapes-trained model on  instance segmentation. Here, we consider only `things' instances during evaluation, and hence we disregard the pixels corresponding to the `stuff' classes. %

\paragraph{Metrics} We evaluate instance segmentation using the standard metrics \cite{cordts2016cityscapes}: 1) \emph{Average Precision} (AP) computes true positives using a number of overlap thresholds, averages over these thresholds, and then averages over classes; 2) AP50 computes average precision with an overlap threshold of $0.5$ and then averages across classes.

\paragraph{Baselines} We compare against the baselines presented by Graber \etal~\cite{graber2021panoptic}. F2F is introduced by Luc \etal~\cite{luc2018predicting} and predicts the features of a future scene using a convolutional model. It then obtains instances by passing these features through MaskRCNN  heads. IndRNN-Stack is the independent RNN and stacking model by Graber \etal~\cite{graber2021panoptic}. PFA, introduced by Lin \etal~\cite{lin2021predictive}, compresses input feature pyramids into a low-resolution feature map for forecasting. %

\paragraph{Results}
The results for this task are presented in \cref{tab:inst_val}. We outperform F2F and IndRNN-Stack in the mid-term setting but PFA performs better. 
This is to be expected because PFA was directly trained on instance segmentation while we directly apply the model trained on panoptic segmentation, \ie, we don't retrain our model specifically for instance segmentation.

\begin{table}[t]
\centering
\begin{tabular}{lcccc}
\toprule
&\multicolumn{2}{c}{Short term: $\Delta t = 3$} & \multicolumn{2}{c}{Mid term: $\Delta t = 9$} \\
& AP & AP50 & AP & AP50 \\
\midrule
Oracle & $34.6$ & $57.4$ & $34.6$ & $57.4$\\
Last seen frame & $8.9$ & $21.3$ & $1.7$ & $6.6$\\
\midrule
F2F \cite{luc2018predicting} & $19.4$ & $\textbf{39.9}$ & $7.7$ & $19.4$ \\

IndRNN-Stack \cite{graber2021panoptic} & $17.8$ & $38.4$ & $10.0$ & $22.3$\\
PFA \cite{lin2021predictive} & $\textbf{24.9}$ & $\textbf{48.7}$ & $\textbf{14.8}$ & $\textbf{30.5}$ \\
\textbf{Ours} & $19.9$ & $39.9$ & $11.2$ & $25.2$ \\
\bottomrule
\end{tabular}
\caption{\textbf{Instance segmentation forecasting on the Cityscapes Validation dataset.} Higher is better for all metrics.}
\label{tab:inst_val}
\end{table}

\begin{table}[t]
  \centering
  \begin{tabular}{lcccc}
\toprule
& \multicolumn{2}{c}{Short term: $\Delta t = 3$} & \multicolumn{2}{c}{Mid term: $\Delta t = 9$} \\
Accuracy (mIoU) & All & MO & All & MO \\
\midrule
Oracle                 &  $80.6$ & $81.7$ & $80.6$ & $81.7$ \\
\midrule
Copy last   & $59.1$  &  $55.0$ & $42.4$ & $33.4$ \\
Bayesian S2S \cite{bhattacharyya2019bayesian}        &  $65.1$ & / & $51.2$ & / \\
DeformF2F \cite{vsaric2019single}          &  $65.5$ & $63.8$ & $53.6$ & $49.9$ \\
LSTM M2M \cite{terwilliger2019recurrent}           &  $67.1$ & $65.1$ & $51.5$ & $46.3$ \\
F2MF \cite{saric2020warp}     &  $69.6$ & $67.7$ & $57.9$ & $54.6$ \\
IndRNN-Stack \cite{graber2021panoptic}          &  $67.6$ & $60.8$ & $58.1$ & $52.1$\\
PFA \cite{lin2021predictive} & $\textbf{71.1}$ & $\textbf{69.2}$ & $\textbf{60.3}$ & $\textbf{56.7}$ \\
\textbf{Ours} & $67.9$ & $61.2$ & $58.1$ & $51.7$ \\
\bottomrule
  \end{tabular}
  \caption{\textbf{Semantic forecasting results on the Cityscapes validation dataset}.  Baseline numbers, besides oracle and copy last, are from \cite{saric2020warp}. Higher is better for all metrics. Our model exploits stereo and odometry, which are provided by typical autonomous vehicle setups and are included in Cityscapes.}
  \label{tab:semantic}
 \end{table}

\section{Cityscapes Semantic Segmentation}
\label{app:semseg}
Following prior work~\cite{graber2021panoptic}, we also evaluate our model on  semantic segmentation forecasting. In this context, we do not care about specific instances. Hence, for each pixel, we discard all predicted identity information.

\noindent{\textbf{Metrics.}} Semantic segmentation forecasting is evaluated using the standard \emph{intersection over union} (IoU) metric computed between predictions and ground truth per class and averaged over classes. IoU (MO), meanwhile, computes an average IoU over `things' classes only.

\noindent{\textbf{Baselines.}} Many of the baselines operate by predicting the features of a future scene 
\cite{bhattacharyya2019bayesian, vsaric2019single, saric2020warp,lin2021predictive}. 
LSTM M2M \cite{terwilliger2019recurrent} warps input semantics using a predicted optical flow between the most recent frame and the target frame. Note that these approaches do not use depth inputs, and all except Bayesian S2S~\cite{bhattacharyya2019bayesian} do not use egomotion as input.

\noindent{\textbf{Results.}}
The results for this task are given in \cref{tab:semantic}. We outperform IndRNN-Stack by a small margin in the short-term setting, and have comparable results in the mid-term setting. We additionally outperform most other baselines. Note that this metric does not care about the boundaries between individual instances and hence weights some types of errors differently than the other metrics we use. These other metrics more properly evaluate whether specific instances are localized in the correct places, which we argue better captures the goals of forecasting. 
Note that PFA is directly trained on semantic segmentation forecasting while our approach was trained on forecasting of panoptic segmentations.

\section{Additional Cityscapes Visualizations}
\label{app:additionalviz}
\begin{figure*}[t]
\centering
\footnotesize
\setlength\tabcolsep{0.5pt}
\renewcommand{\arraystretch}{0.5}
\begin{tabular}{cccc}
\rotatebox{90}{\hspace{0.5cm}IndRNN-Stack} & \includegraphics[width=0.32\textwidth]{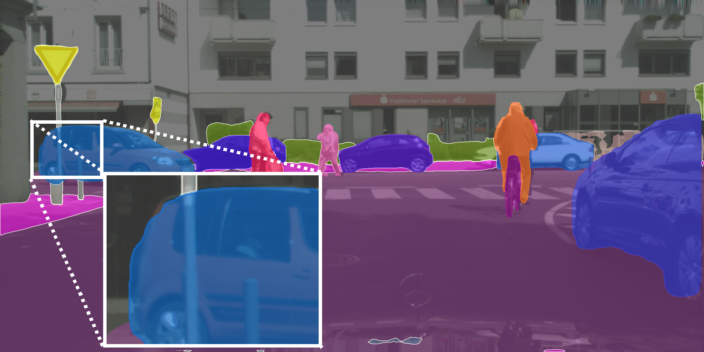}& \includegraphics[width=0.32\textwidth]{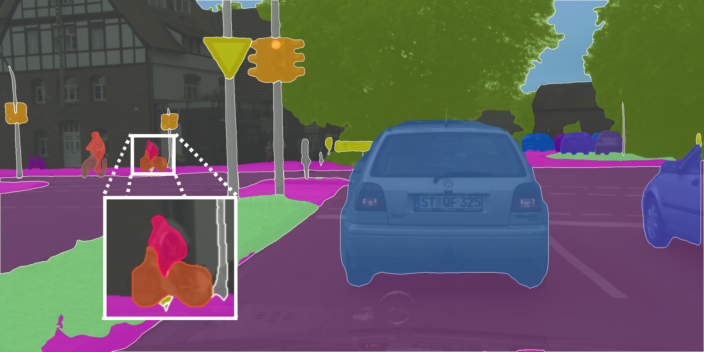} & \includegraphics[width=0.32\textwidth]{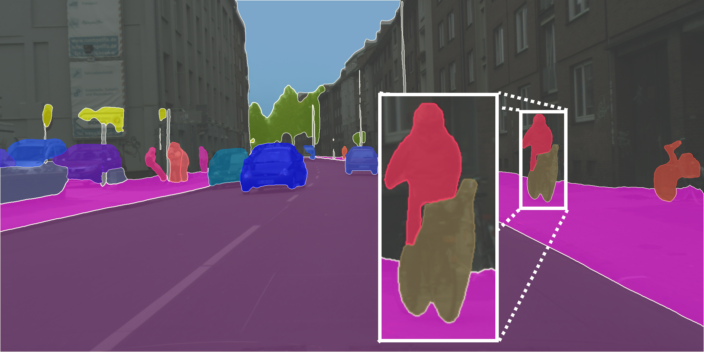} \\ 
\rotatebox{90}{\hspace{1.0cm}\textbf{Ours}} & \includegraphics[width=0.32\textwidth]{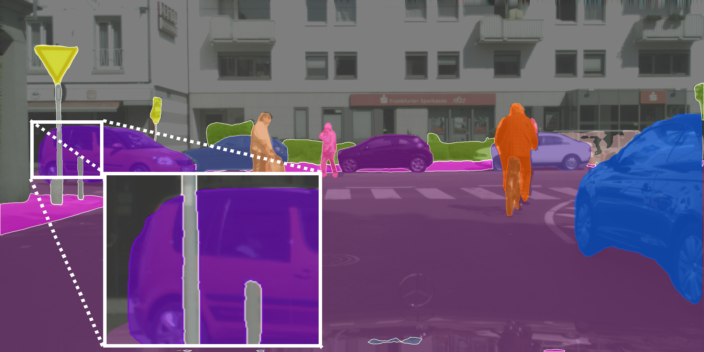}& \includegraphics[width=0.32\textwidth]{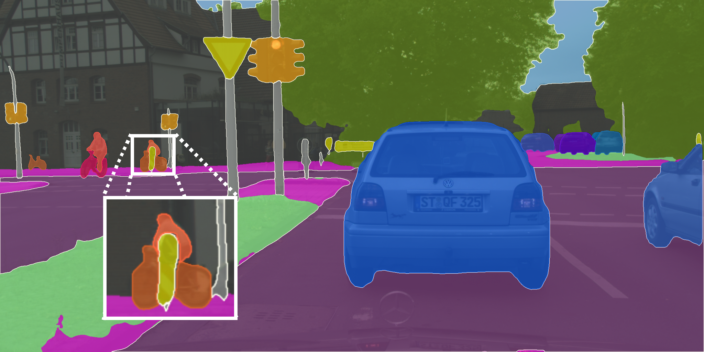} & \includegraphics[width=0.32\textwidth]{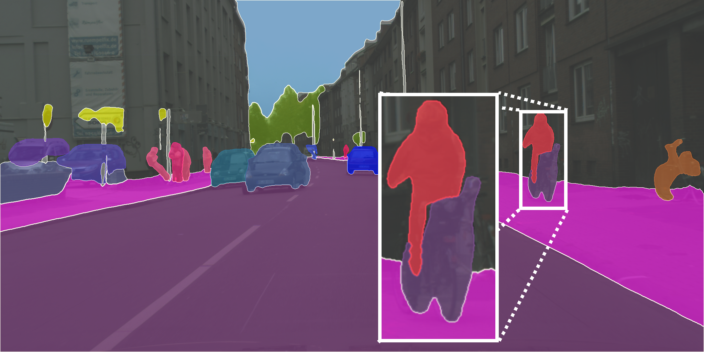}
\end{tabular}
\vspace{-0.3cm}
\caption{Short-term panoptic segmentation forecasts on Cityscapes.}
\label{fig:panoptic_viz_short}
\vspace{-0.4cm}
\end{figure*}

\begin{figure*}[t]
\centering
\footnotesize
\setlength\tabcolsep{0.5pt}
\renewcommand{\arraystretch}{0.5}
\begin{tabular}{cccc}
\rotatebox{90}{\hspace{0.5cm}IndRNN-Stack} & \includegraphics[width=0.32\textwidth]{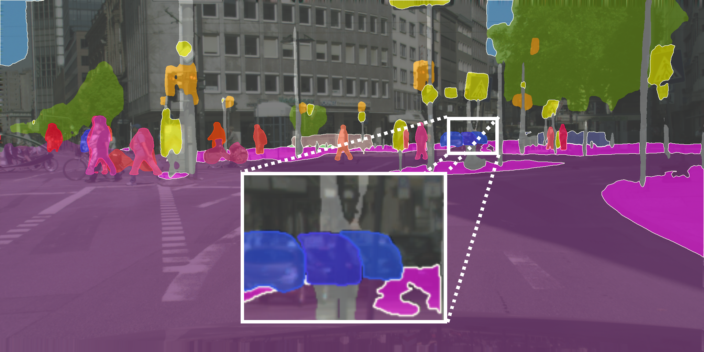}& \includegraphics[width=0.32\textwidth]{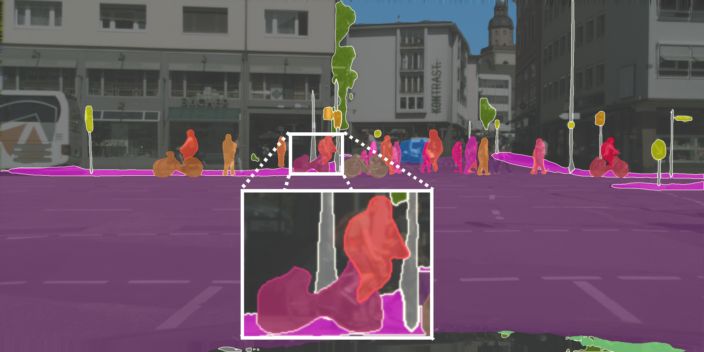} & \includegraphics[width=0.32\textwidth]{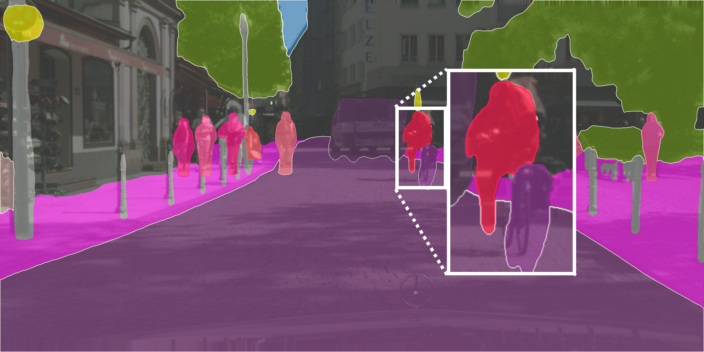} \\ 
\rotatebox{90}{\hspace{1.0cm}\textbf{Ours}} & \includegraphics[width=0.32\textwidth]{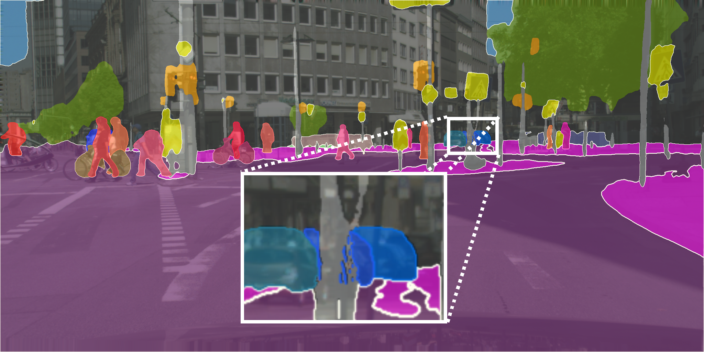}& \includegraphics[width=0.32\textwidth]{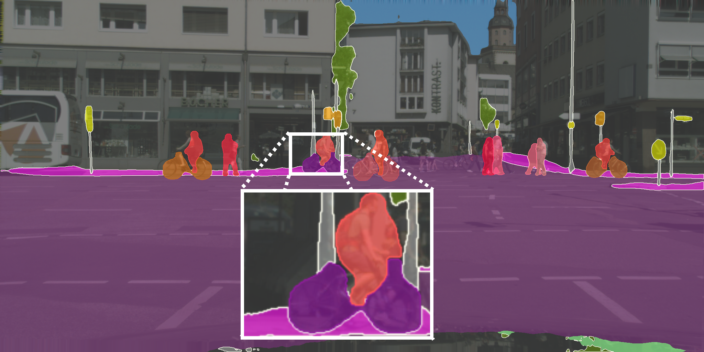} & \includegraphics[width=0.32\textwidth]{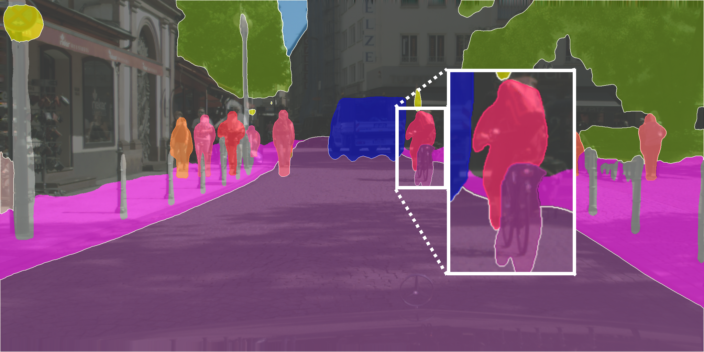} \\
\vspace{0.5cm}
\\
\rotatebox{90}{\hspace{0.5cm}IndRNN-Stack} & \includegraphics[width=0.32\textwidth]{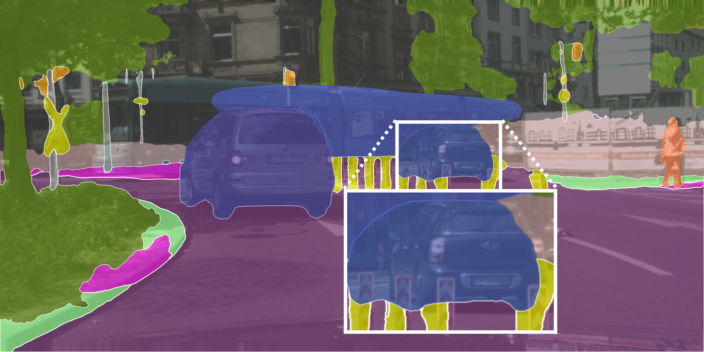}& \includegraphics[width=0.32\textwidth]{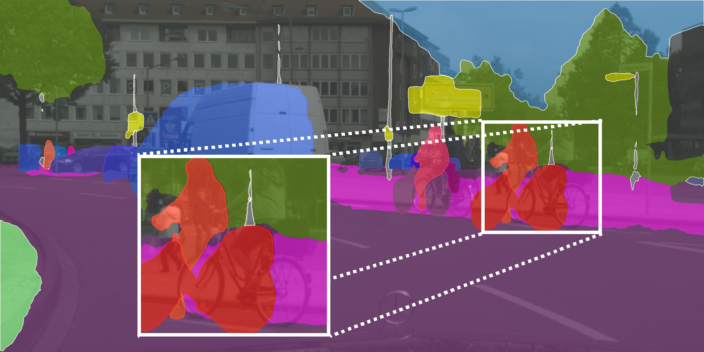} & \includegraphics[width=0.32\textwidth]{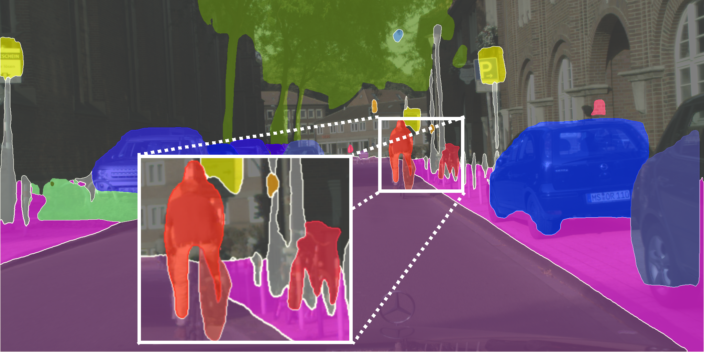} \\ 
\rotatebox{90}{\hspace{1.0cm}\textbf{Ours}} & \includegraphics[width=0.32\textwidth]{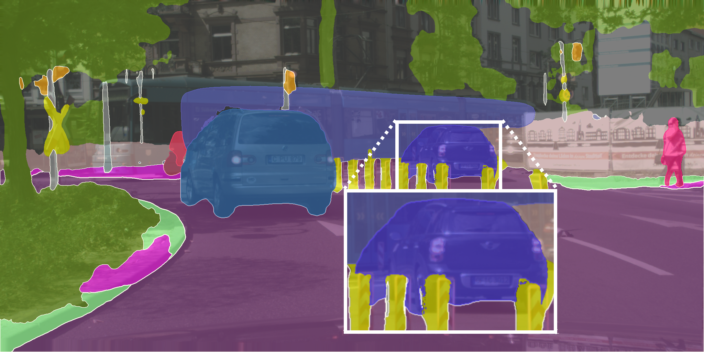}& \includegraphics[width=0.32\textwidth]{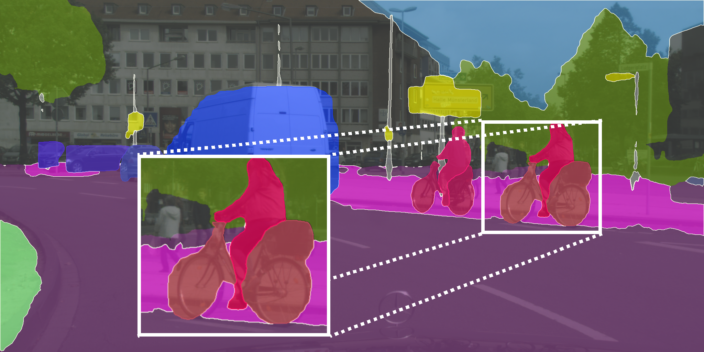} & \includegraphics[width=0.32\textwidth]{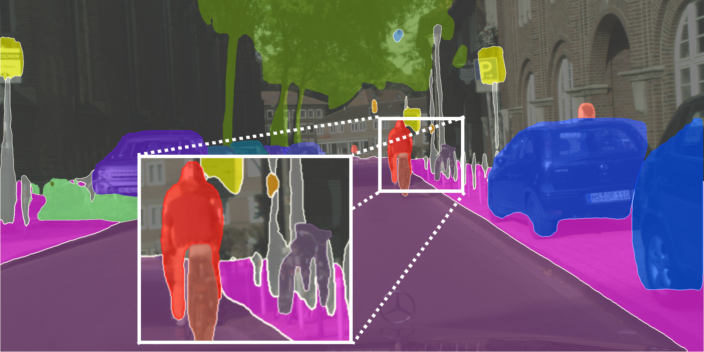} \\
\vspace{0.5cm}
\\
\rotatebox{90}{\hspace{0.5cm}IndRNN-Stack} & \includegraphics[width=0.32\textwidth]{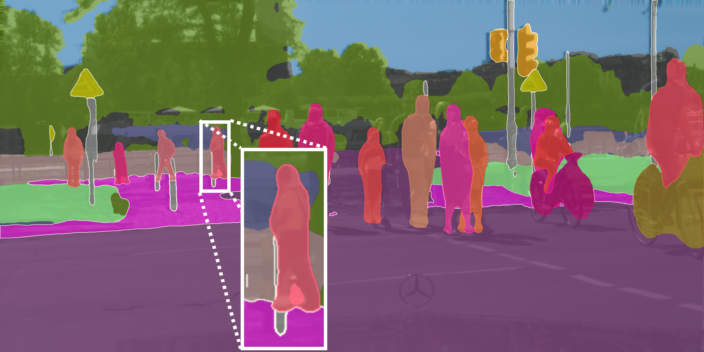} & \includegraphics[width=0.32\textwidth]{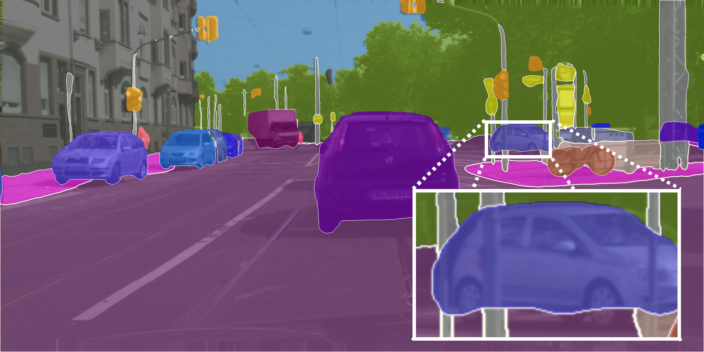} & \includegraphics[width=0.32\textwidth]{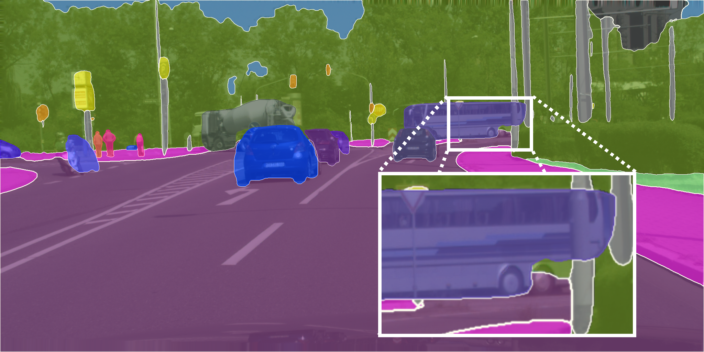} \\
\rotatebox{90}{\hspace{1.0cm}\textbf{Ours}} & \includegraphics[width=0.32\textwidth]{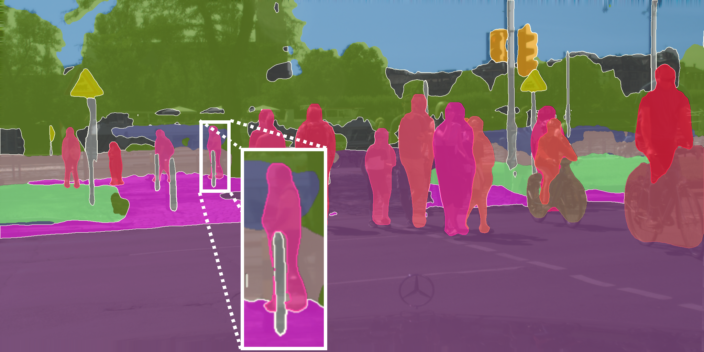} & \includegraphics[width=0.32\textwidth]{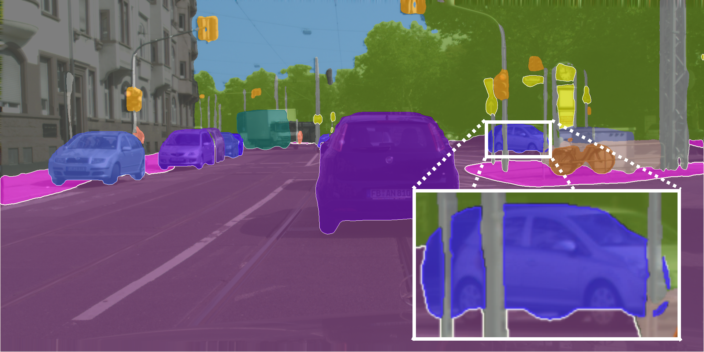} & \includegraphics[width=0.32\textwidth]{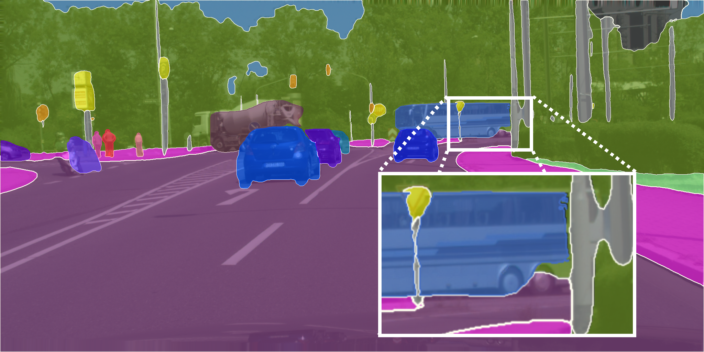}\\
\end{tabular}
\vspace{-0.3cm}
\caption{Additional mid-term visualizations.}
\label{fig:viz_mid_supp_1}
\vspace{-0.4cm}
\end{figure*}

\begin{figure*}[t]
\centering
\footnotesize
\setlength\tabcolsep{0.5pt}
\renewcommand{\arraystretch}{0.5}
\begin{tabular}{cccc}
\rotatebox{90}{\hspace{0.5cm}IndRNN-Stack} & \includegraphics[width=0.32\textwidth]{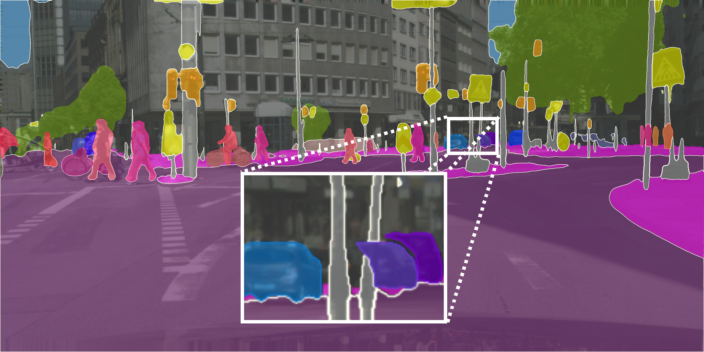}& \includegraphics[width=0.32\textwidth]{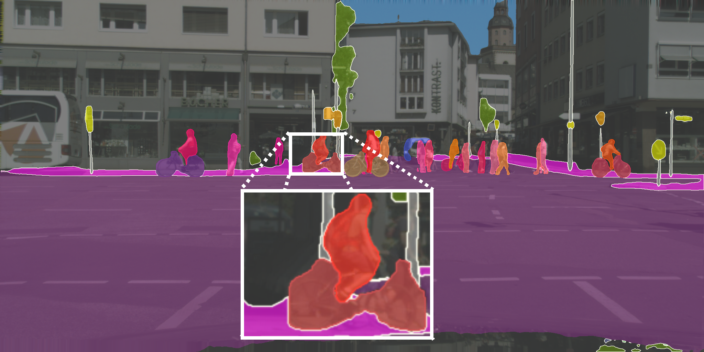} & \includegraphics[width=0.32\textwidth]{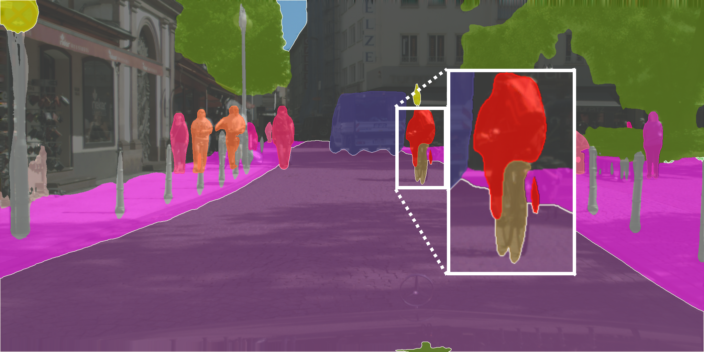} \\ 
\rotatebox{90}{\hspace{1.0cm}\textbf{Ours}} & \includegraphics[width=0.32\textwidth]{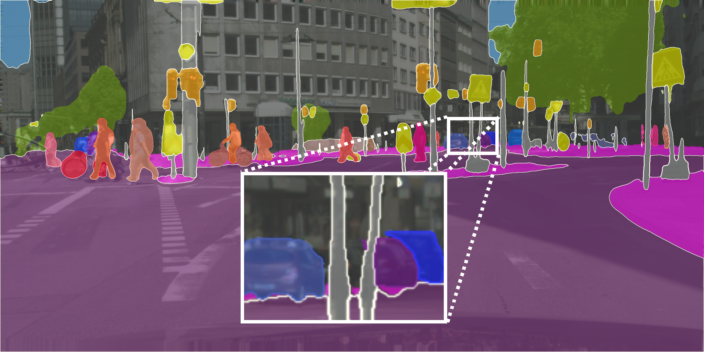}& \includegraphics[width=0.32\textwidth]{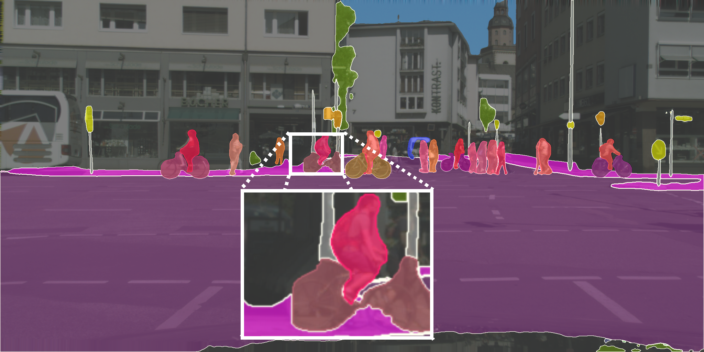} & \includegraphics[width=0.32\textwidth]{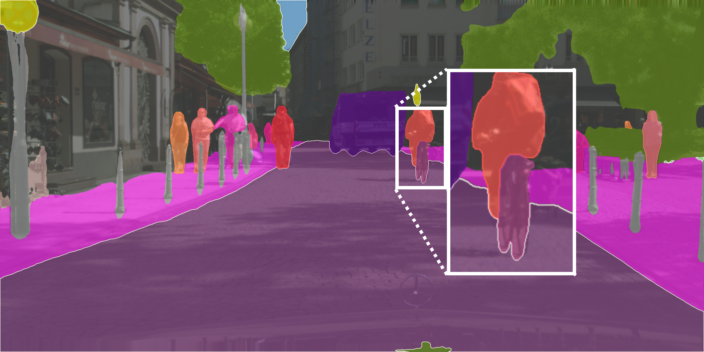} \\
\vspace{0.5cm}
\\
\rotatebox{90}{\hspace{0.5cm}IndRNN-Stack} & \includegraphics[width=0.32\textwidth]{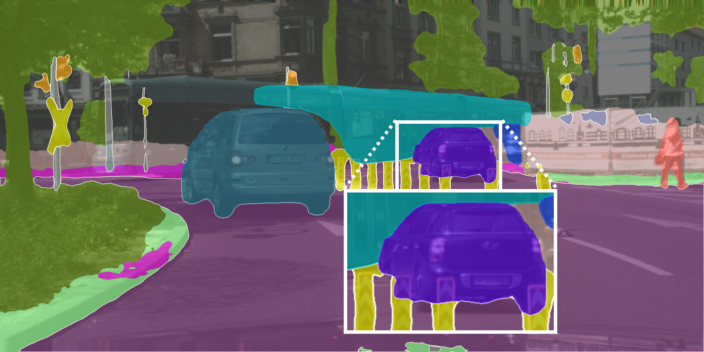}& \includegraphics[width=0.32\textwidth]{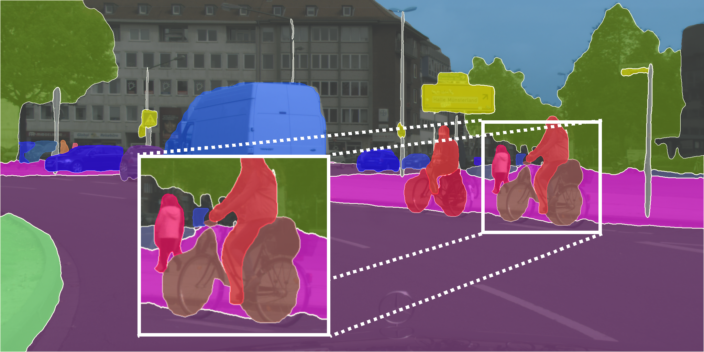} & \includegraphics[width=0.32\textwidth]{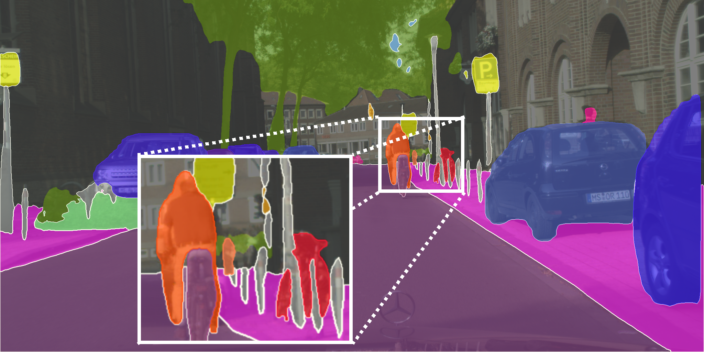} \\ 
\rotatebox{90}{\hspace{1.0cm}\textbf{Ours}} & \includegraphics[width=0.32\textwidth]{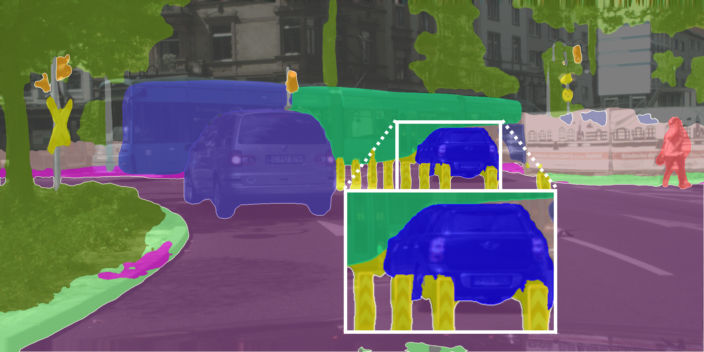}& \includegraphics[width=0.32\textwidth]{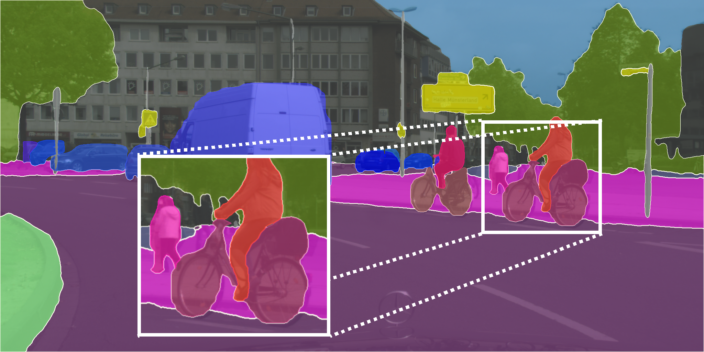} & \includegraphics[width=0.32\textwidth]{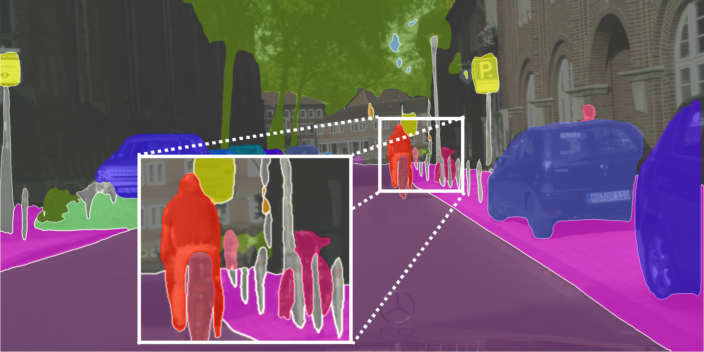} \\
\vspace{0.5cm}
\\
\rotatebox{90}{\hspace{0.5cm}IndRNN-Stack} & \includegraphics[width=0.32\textwidth]{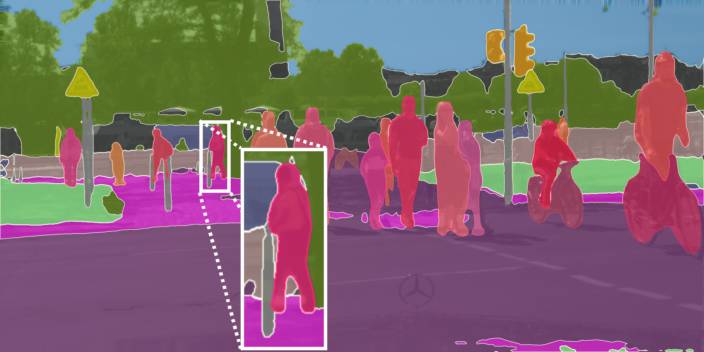} & \includegraphics[width=0.32\textwidth]{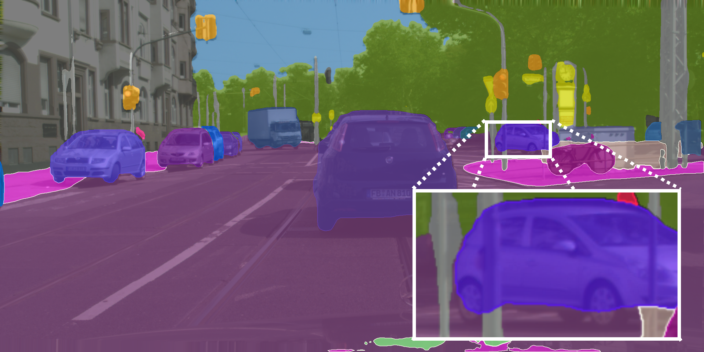} & \includegraphics[width=0.32\textwidth]{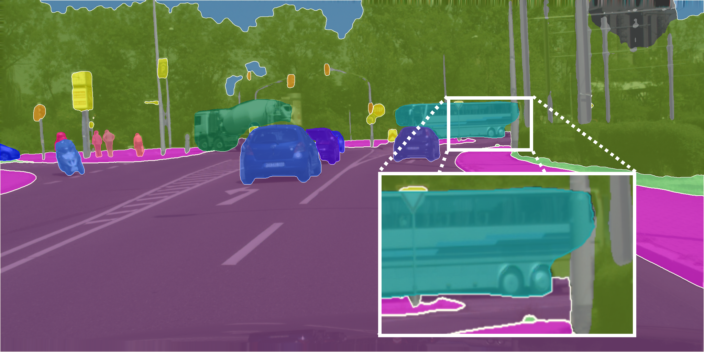} \\
\rotatebox{90}{\hspace{1.0cm}\textbf{Ours}} & \includegraphics[width=0.32\textwidth]{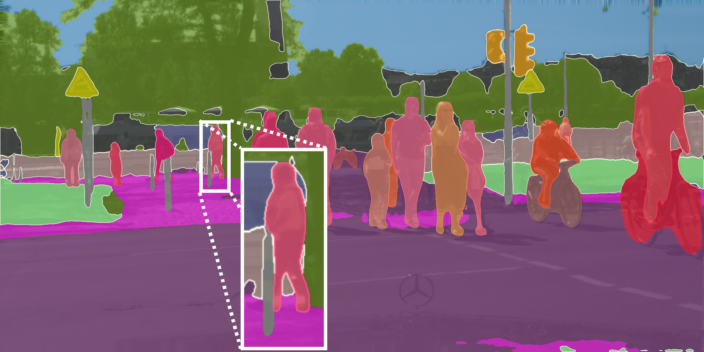} & \includegraphics[width=0.32\textwidth]{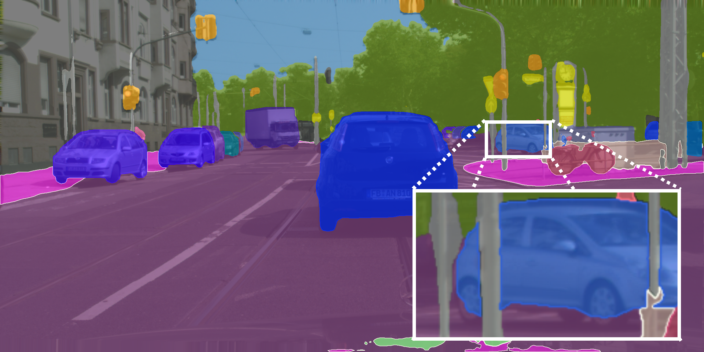} & \includegraphics[width=0.32\textwidth]{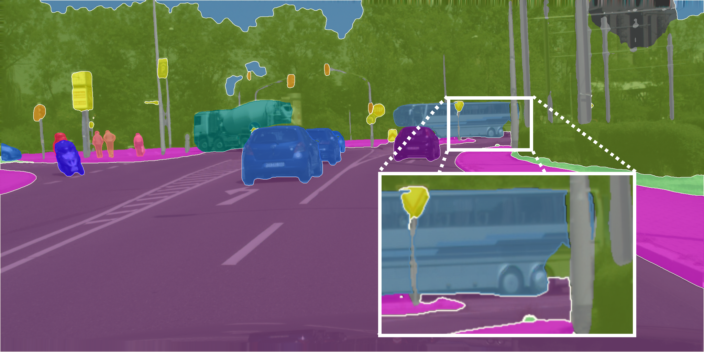}\\
\end{tabular}
\vspace{-0.3cm}
\caption{Additional short-term visualizations.}
\label{fig:viz_short_supp_1}
\vspace{-0.4cm}
\end{figure*}

\cref{fig:panoptic_viz_short} presents a visual comparison between our approach and IndRNN-Stack for the short-term setting for the sequences which were shown for the mid-term setting in \cref{fig:panoptic_viz}. We present additional visualizations for the mid-term setting in \cref{fig:viz_mid_supp_1} and for the short term setting in \cref{fig:viz_short_supp_1}.

\section{Additional Cityscapes Metrics}
\label{app:full_metrics}
\begin{table*}[t]
    \vspace{-0.2cm}
      \centering
      \begin{tabular}{l ccc ccc ccc}
    \toprule

    & \multicolumn{3}{c}{All} & \multicolumn{3}{c}{Things} & \multicolumn{3}{c}{Stuff}\\
    & PQ & SQ & RQ & PQ & SQ & RQ & PQ & SQ & RQ \\
    \midrule
    Flow & $25.6$ & $70.1$ & $34.0$ & $12.4$ & $66.3$ & $18.1$ & $35.3$ & $72.9$ & $45.5$\\
    Hybrid \cite{terwilliger2019recurrent} (bg) and \cite{luc2018predicting} (fg) & $29.4$ & $69.8$ & $38.5$ & $18.0$ & $67.2$ & $25.7$& $37.6$ & $71.6$ & $47.8$\\
    IndRNN-Stack & $35.7$ & $72.0$ & $46.5$ & $24.0$ & $69.0$ & $33.7$ & $44.2 $ & $74.2$ & $55.8$\\
    \textbf{Ours} & $\textbf{36.9}$ & $\textbf{72.7}$ & $\textbf{48.0}$ & $\textbf{26.7}$ & $\textbf{70.3}$ & $\textbf{37.0}$ & $\textbf{44.4}$ & $\textbf{74.4}$ & $\textbf{55.9}$ \\
    \bottomrule
      \end{tabular}
      \caption{\textbf{Panoptic segmentation forecasting evaluated on the Cityscapes test set, mid-term.}  
      Higher is better for all metrics.}
      \label{tab:panoptic_test}
 \end{table*}
 
 \begin{table}\centering
 \begin{minipage}[t]{0.47\textwidth}\centering
\begin{tabular}{lcc}
\toprule
 & AP & AP50 \\
\midrule
F2F \cite{luc2018predicting} & $6.7$ & $17.5$ \\
IndRNN-Stack & $8.4$ & $19.8$\\
\textbf{Ours} & $\textbf{9.9}$ & $\textbf{20.7}$ \\
\bottomrule
\end{tabular}
\caption{\textbf{Instance segmentation forecasting on the Cityscapes Test dataset, mid-term.} Higher is better for all metrics.}
\label{tab:inst_test}
 \end{minipage}
 \hfill
 \begin{minipage}[t]{0.47\textwidth}\centering 
 \begin{tabular}{lcc}
\toprule
Accuracy (mIoU) & All & MO \\
\midrule
F2MF \cite{saric2020warp}$^*$ & $59.1$ & $56.3$\\
IndRNN-Stack & $57.7$ & $48.8$  \\
\textbf{Ours} & $58.3$ & $50.0$ \\
\bottomrule
  \end{tabular}
  \vspace{1pt}
  \caption{\textbf{Semantic segmentation forecasting results on the Cityscapes test dataset.}  Baseline numbers,  are from \cite{saric2020warp}; the * indicates training on both train and validation data. Higher is better for all metrics.}
  \label{tab:test_semantic}
 \end{minipage}
\end{table}

\crefrange{tab:panoptic_test}{tab:test_semantic} present metrics computed for the Cityscapes test dataset using the mid-term setting for panoptic, instance, and semantic segmentation forecasting, respectively. We outperform all other approaches for panoptic and instance segmentation forecasting on the test data. On semantic segmentation, we outperform IndRNN-Stack on the test data, whereas F2MF \cite{saric2020warp} outperforms our approach. However, note that the F2MF model used for test evaluation was trained on both the training and validation datasets, while the other models were trained only on the training data.

\crefrange{tab:per_class_pq_short}{tab:per_class_rq_mid} contain the per-class breakdown of all panoptic segmentation metrics shown in \cref{tab:panoptic}. The results shown in \cref{tab:panoptic} consist of the average of these metrics taken over the values obtained for each individual class. Our model is better on average for every metric than all other approaches, and it is additionally better than prior approaches for every metric for most classes.

\section{Code Details}
\label{app:code}
All models are implemented using PyTorch v. $1.10.0$ \footnote{\url{https://pytorch.org/}}, which is made available for use with a custom BSD-style license.\footnote{\url{https://github.com/pytorch/pytorch/blob/v1.10.0/LICENSE}} We additionally use the Detectron2 framework (version 0.4.1)\footnote{\url{https://github.com/facebookresearch/detectron2}}, which is released under the Apache 2.0 license.\footnote{\url{https://github.com/facebookresearch/detectron2/blob/v0.4.1/LICENSE}} 
Code implementing our models and experiments can be found at \url{https://github.com/cgraber/psf-diffattn}.

\section{Potential Negative Societal Impact}
\label{app:impact}
One of the primary applications of this work is to better enhance the ability of autonomous agents to anticipate the future and respond appropriately to a dynamic environment. In this context, problems can arise if an agent makes a decision based on a faulty prediction -- for example, if a self-driving car does not anticipate a pedestrian stepping into the street, it could unintentionally hurt the pedestrian if they step out in front of the car. For such a system, the consequence of prediction errors can be injury or death. It is thus critical that appropriate care be taken before deployment of such a system to ensure that not only are prediction errors sufficiently low across a variety of environments but also that proper failsafes are put in place to minimize the negative consequences of acting upon a misprediction. 

\begin{table*}[t]
    \centering
    \newcommand*\rot{\rotatebox{90}}
    
    \resizebox{1.0\textwidth}{!}{
    \begin{tabular}{l|ccccccccccccccccccc|c}
    \toprule
        & \rot{road} & \rot{sidewalk} & \rot{building} & \rot{wall} & \rot{fence} & \rot{pole} & \rot{traffic light} & \rot{traffic sign} & \rot{vegetation} & \rot{terrain} & \rot{sky} & \rot{person} & \rot{rider} & \rot{car} & \rot{truck} & \rot{bus} & \rot{train} & \rot{motorcycle} & \rot{bicycle} & \rot{mean} \\
        \midrule
        \textbf{Deeplab (Oracle)$\dagger$} & $97.9$ & $78.2$ & $88.5$ & $29.4$ & $38.9$ & $60.0$ & $55.6$ & $74.5$ & $89.5$ & $36.1$ & $87.9$ & $50.8$ & $46.4$ & $67.3$ & $51.5$ & $66.6$ & $37.8$ & $44.2$ & $44.1$ & $60.3$\\
        \midrule
        \textbf{Deeplab (Last seen frame)} & $94.3$ & $52.4$ & $71.1$ & $11.3$ & $19.4$ & $6.1$ & $12.9$ & $15.0$ & $72.1$ & $16.9$ & $72.7$ & $10.3$ & $8.0$ & $29.6$ & $35.1$ & $51.7$ & $24.2$ & $9.8$ & $7.9$ & $32.7$ \\
        Flow & $95.6$ & $61.5$ & $79.8$ & $17.3$ & $\textbf{28.6}$ & $8.7$ & $26.2$ & $36.8$ & $80.7$ & $\textbf{26.9}$ & $79.7$ & $21.0$ & $14.0$ & $43.4$ & $40.6$ & $56.8$ & $26.7$ & $23.2$ & $18.7$ & $41.4$\\
        Hybrid \cite{terwilliger2019recurrent} (bg) and \cite{luc2018predicting} (fg) & $\textbf{96.2}$ & $63.4$ & $81.4$ & $23.1$ & $23.7$ & $7.1$ & $19.1$ & $36.9$ & $82.3$ & $20.3$ & $79.8$ & $26.8$ & $21.8$ & $46.4$ & $\textbf{42.2}$ & $60.0$ & $41.4$ & $25.6$ & $22.5$ & $43.2$\\
        IndRNN-Stack & $\textbf{96.2}$ & $66.1$ & $83.5$ & $\textbf{26.1}$ & $27.4$ & $31.7$ & $\textbf{37.0}$ & $49.9$ & $\textbf{84.8}$ & $26.1$ & $82.0$ & $31.8$ & $31.5$ & $48.8$ & $\textbf{42.2}$ & $61.2$ & $47.0$ & $31.4$ & $27.3$ & $49.0$\\
        \textbf{Ours} & $\textbf{96.2}$ & $\textbf{66.3}$ & $\textbf{83.8}$ & $25.9$ & $27.4$ & $\textbf{34.4}$ & $\textbf{37.0}$ & $\textbf{50.3}$ & $\textbf{84.8}$ & $26.5$ & $\textbf{82.1}$ & $\textbf{34.9}$ & $\textbf{36.7}$ & $\textbf{51.2}$ & $41.2$ & $\textbf{63.1}$ & $\textbf{47.6}$ & $\textbf{32.4}$ & $\textbf{32.0}$ & $\textbf{50.2}$\\
        \bottomrule
    \end{tabular}}
    
    \vspace{5pt}
    \caption{\textbf{Per-class results for Panoptic Quality on Cityscapes validation dataset (short-term).} }
    \label{tab:per_class_pq_short}
\end{table*}
\begin{table*}[t]
    \centering
    \newcommand*\rot{\rotatebox{90}}
    
    \resizebox{1.0\textwidth}{!}{
    \begin{tabular}{l|ccccccccccccccccccc|c}
    \toprule
        & \rot{road} & \rot{sidewalk} & \rot{building} & \rot{wall} & \rot{fence} & \rot{pole} & \rot{traffic light} & \rot{traffic sign} & \rot{vegetation} & \rot{terrain} & \rot{sky} & \rot{person} & \rot{rider} & \rot{car} & \rot{truck} & \rot{bus} & \rot{train} & \rot{motorcycle} & \rot{bicycle} & \rot{mean} \\
        \midrule
        \textbf{Deeplab (Oracle)$\dagger$} & $97.9$ & $78.2$ & $88.5$ & $29.4$ & $38.9$ & $60.0$ & $55.6$ & $74.5$ & $89.5$ & $36.1$ & $87.9$ & $50.8$ & $46.4$ & $67.3$ & $51.5$ & $66.6$ & $37.8$ & $44.2$ & $44.1$ & $60.3$ \\
        \midrule
        \textbf{Deeplab (Last seen frame)} & $90.4$ & $32.5$ & $57.6$ & $7.6$ & $10.6$ & $4.6$ & $8.9$ & $7.4$ & $55.1$ & $8.8$ & $57.3$ & $5.3$ & $2.5$ & $13.2$ & $19.2$ & $27.3$ & $10.1$ & $4.7$ & $3.0$ & $22.4$  \\
        Flow & $90.5$ & $35.8$ & $66.2$ & $7.7$ & $15.0$ & $4.6$ & $11.9$ & $11.1$ & $65.6$ & $11.6$ & $64.4$ & $5.9$ & $2.5$ & $19.0$ & $21.5$ & $27.7$ & $13.5$ & $11.8$ & $5.3$ & $25.9$\\
        Hybrid \cite{terwilliger2019recurrent} (bg) and \cite{luc2018predicting} (fg) & $93.2$ & $44.9$ & $70.5$ & $12.4$ & $14.8$ & $1.2$ & $8.0$ & $10.8$ & $69.7$ & $13.9$ & $67.2$ & $8.0$ & $4.5$ & $27.3$ & $33.5$ & $41.7$ & $27.9$ & $8.3$ & $6.1$ & $29.7$\\
        IndRNN-Stack & $\textbf{93.9}$ & $50.8$ & $76.4$ & $\textbf{18.2}$ & $19.9$ & $8.7$ & $18.7$ & $28.5$ & $\textbf{77.0}$ & $\textbf{18.6}$ & $\textbf{72.7}$ & $16.2$ & $12.0$ & $33.3$ & $36.1$ & $53.0$ & $\textbf{29.8}$ & $14.1$ & $12.6$ & $36.3$\\
        \textbf{Ours} & $\textbf{93.9}$ & $\textbf{50.9}$ & $\textbf{76.5}$ & $18.1$ & $\textbf{20.8}$ & $\textbf{9.3}$ & $\textbf{18.8}$ & $\textbf{28.6}$ & $\textbf{77.0}$ & $\textbf{18.6}$ & $\textbf{72.7}$ & $\textbf{19.9}$ & $\textbf{14.6}$ & $\textbf{39.5}$ & $\textbf{38.8}$ & $\textbf{56.9}$ & $26.2$ & $\textbf{18.6}$ & $\textbf{14.5}$ & $\textbf{37.6}$ \\
        \bottomrule
    \end{tabular}}
    
    \vspace{5pt}
    \caption{\textbf{Per-class results for Panoptic Quality on Cityscapes validation dataset (mid-term).} }
    \label{tab:per_class_pq_mid}
\end{table*}
\begin{table*}[t]
    \centering
    \newcommand*\rot{\rotatebox{90}}
    
    \resizebox{1.0\textwidth}{!}{
    \begin{tabular}{l|ccccccccccccccccccc|c}
    \toprule
        & \rot{road} & \rot{sidewalk} & \rot{building} & \rot{wall} & \rot{fence} & \rot{pole} & \rot{traffic light} & \rot{traffic sign} & \rot{vegetation} & \rot{terrain} & \rot{sky} & \rot{person} & \rot{rider} & \rot{car} & \rot{truck} & \rot{bus} & \rot{train} & \rot{motorcycle} & \rot{bicycle} & \rot{mean} \\
        \midrule
        \textbf{Deeplab (Oracle)$\dagger$} & $98.0$ & $85.6$ & $90.5$ & $74.3$ & $74.8$ & $69.7$ & $73.5$ & $80.1$ & $90.9$ & $75.7$ & $92.6$ & $76.0$ & $70.8$ & $84.2$ & $88.4$ & $90.8$ & $87.6$ & $73.8$ & $72.1$ & $81.5$ \\
        \midrule
        \textbf{Deeplab (Last seen frame)} & $94.4$ & $71.5$ & $78.8$ & $65.4$ & $65.6$ & $\textbf{67.0}$ & $\textbf{68.3}$ & $67.4$ & $77.8$ & $67.7$ & $83.0$ & $64.4$ & $60.1$ & $69.2$ & $74.7$ & $76.7$ & $75.7$ & $62.7$ & $63.4$ & $71.3$ \\
        Flow & $95.6$ & $76.0$ & $83.2$ & $68.5$ & $68.3$ & $65.0$ & $65.9$ & $67.3$ & $83.4$ & $69.1$ & $86.6$ & $65.6$ & $61.4$ & $75.8$ & $77.5$ & $80.0$ & $74.1$ & $66.1$ & $64.4$ & $73.4$\\
        Hybrid \cite{terwilliger2019recurrent} (bg) and \cite{luc2018predicting} (fg) & $\textbf{96.3}$ & $\textbf{77.2}$ & $84.9$ & $70.0$ & $69.0$ & $59.5$ & $63.6$ & $65.9$ & $84.6$ & $70.8$ & $86.5$ & $66.8$ & $61.9$ & $77.2$ & $80.3$ & $83.1$ & $\textbf{80.5}$ & $65.6$ & $63.8$ & $74.1$\\
        IndRNN-Stack & $\textbf{96.3}$ & $77.0$ & $86.3$ & $71.1$ & $69.4$ & $61.4$ & $65.4$ & $70.6$ & $\textbf{86.6}$ & $\textbf{71.3}$ & $\textbf{88.3}$ & $67.7$ & $63.8$ & $77.7$ & $81.4$ & $81.4$ & $74.8$ & $67.6$ & $65.8$ & $74.9$\\
        \textbf{Ours} & $\textbf{96.3}$ & $77.1$ & $\textbf{86.4}$ & $\textbf{71.4}$ & $\textbf{69.8}$ & $61.7$ & $65.4$ & $\textbf{70.8}$ & $\textbf{86.6}$ & $70.9$ & $\textbf{88.3}$ & $\textbf{69.4}$ & $\textbf{66.4}$ & $\textbf{78.9}$ & $\textbf{81.7}$ & $\textbf{84.1}$ & $77.9$ & $\textbf{68.1}$ & $\textbf{67.2}$ & $\textbf{75.7}$ \\
        \bottomrule
    \end{tabular}}
    
    \vspace{5pt}
    \caption{\textbf{Per-class results for Segmentation Quality on Cityscapes validation dataset (short-term).} }
    \label{tab:per_class_sq_short}
\end{table*}
\begin{table*}[t]
    \centering
    \newcommand*\rot{\rotatebox{90}}
    
    \resizebox{1.0\textwidth}{!}{
    \begin{tabular}{l|ccccccccccccccccccc|c}
    \toprule
        & \rot{road} & \rot{sidewalk} & \rot{building} & \rot{wall} & \rot{fence} & \rot{pole} & \rot{traffic light} & \rot{traffic sign} & \rot{vegetation} & \rot{terrain} & \rot{sky} & \rot{person} & \rot{rider} & \rot{car} & \rot{truck} & \rot{bus} & \rot{train} & \rot{motorcycle} & \rot{bicycle} & \rot{mean} \\
        \midrule
        \textbf{Deeplab (Oracle)$\dagger$} & $98.0$ & $85.6$ & $90.5$ & $74.3$ & $74.8$ & $69.7$ & $73.5$ & $80.1$ & $90.9$ & $75.7$ & $92.6$ & $76.0$ & $70.8$ & $84.2$ & $88.4$ & $90.8$ & $87.6$ & $73.8$ & $72.1$ & $81.5$ \\
        \midrule
        \textbf{Deeplab (Last seen frame)} & $90.7$ & $68.2$ & $72.6$ & $63.4$ & $62.4$ & $\textbf{66.1}$ & $\textbf{72.7}$ & $\textbf{73.0}$ & $71.2$ & $64.0$ & $77.3$ & $63.7$ & $61.3$ & $66.8$ & $62.9$ & $70.8$ & $74.3$ & $56.4$ & $64.4$ & $68.5$ \\
        Flow & $90.8$ & $68.6$ & $76.0$ & $66.1$ & $64.1$ & $64.1$ & $69.0$ & $67.2$ & $75.0$ & $64.5$ & $78.5$ & $63.5$ & $60.4$ & $69.1$ & $70.2$ & $74.3$ & $\textbf{75.8}$ & $60.2$ & $63.0$ & $69.5$\\
        Hybrid \cite{terwilliger2019recurrent} (bg) and \cite{luc2018predicting} (fg) & $93.3$ & $69.7$ & $77.9$ & $66.6$ & $65.3$ & $59.9$ & $62.9$ & $61.9$ & $76.9$ & $65.1$ & $79.6$ & $63.7$ & $58.4$ & $71.5$ & $72.6$ & $72.2$ & $73.7$ & $62.1$ & $60.6$ & $69.1$\\
        IndRNN-Stack & $94.1$ & $71.3$ & $81.5$ & $68.4$ & $\textbf{66.8}$ & $59.0$ & $64.1$ & $65.1$ & $80.9$ & $68.1$ & $\textbf{83.0}$ & $64.3$ & $\textbf{61.4}$ & $73.4$ & $\textbf{76.9}$ & $76.1$ & $74.4$ & $62.5$ & $62.9$ & $71.3$ \\
        \textbf{Ours} & $\textbf{94.2}$ & $\textbf{71.5}$ & $\textbf{81.7}$ & $\textbf{69.1}$ & $66.1$ & $60.1$ & $64.2$ & $65.7$ & $\textbf{81.0}$ & $\textbf{68.5}$ & $\textbf{83.0}$ & $\textbf{64.7}$ & $61.1$ & $\textbf{74.7}$ & $76.5$ & $\textbf{79.4}$ & $67.7$ & $\textbf{63.6}$ & $\textbf{64.7}$ & $\textbf{71.4}$ \\
        \bottomrule
    \end{tabular}}
    
    \vspace{5pt}
    \caption{\textbf{Per-class results for Segmentation Quality on Cityscapes validation dataset (mid-term).} }
    \label{tab:per_class_sq_mid}
\end{table*}
\begin{table*}[t]
    \centering
    \newcommand*\rot{\rotatebox{90}}
    
    \resizebox{1.0\textwidth}{!}{
    \begin{tabular}{l|ccccccccccccccccccc|c}
    \toprule
        & \rot{road} & \rot{sidewalk} & \rot{building} & \rot{wall} & \rot{fence} & \rot{pole} & \rot{traffic light} & \rot{traffic sign} & \rot{vegetation} & \rot{terrain} & \rot{sky} & \rot{person} & \rot{rider} & \rot{car} & \rot{truck} & \rot{bus} & \rot{train} & \rot{motorcycle} & \rot{bicycle} & \rot{mean} \\
        \midrule
        \textbf{Deeplab (Oracle)$\dagger$} & $99.9$ & $91.3$ & $97.8$ & $39.5$ & $52.1$ & $86.1$ & $75.6$ & $93.0$ & $98.5$ & $47.7$ & $94.9$ & $66.9$ & $65.5$ & $80.0$ & $58.2$ & $73.4$ & $43.1$ & $59.9$ & $61.2$ & $72.9$ \\
        \midrule
        \textbf{Deeplab (Last seen frame)} & $\textbf{99.9}$ & $73.4$ & $90.2$ & $17.3$ & $29.7$ & $9.1$ & $18.9$ & $22.2$ & $92.7$ & $25.0$ & $87.6$ & $16.0$ & $13.2$ & $42.8$ & $47.0$ & $67.4$ & $32.0$ & $15.6$ & $12.4$ & $42.7$  \\
        Flow & $\textbf{99.9}$ & $81.0$ & $95.9$ & $25.2$ & $\textbf{41.9}$ & $13.4$ & $39.7$ & $54.7$ & $96.8$ & $\textbf{39.0}$ & $92.0$ & $32.1$ & $22.8$ & $57.3$ & $52.4$ & $71.0$ & $36.0$ & $35.1$ & $29.0$ & $53.4$\\
        Hybrid \cite{terwilliger2019recurrent} (bg) and \cite{luc2018predicting} (fg) & $\textbf{99.9}$ & $82.1$ & $95.8$ & $33.0$ & $34.4$ & $11.9$ & $30.0$ & $56.0$ & $97.3$ & $28.8$ & $92.2$ & $40.1$ & $35.3$ & $60.2$ & $\textbf{52.6}$ & $72.2$ & $51.4$ & $39.1$ & $35.3$ & $55.1$\\
        IndRNN-Stack & $\textbf{99.9}$ & $85.8$ & $96.7$ & $\textbf{36.7}$ & $39.4$ & $51.6$ & $\textbf{56.6}$ & $70.7$ & $\textbf{97.9}$ & $36.6$ & $\textbf{92.9}$ & $47.0$ & $49.3$ & $62.9$ & $51.9$ & $\textbf{75.2}$ & $\textbf{62.9}$ & $46.3$ & $41.5$ & $63.3$\\
        \textbf{Ours} & $\textbf{99.9}$ & $\textbf{86.0}$ & $\textbf{97.0}$ & $36.2$ & $39.3$ & $\textbf{55.7}$ & $\textbf{56.6}$ & $\textbf{71.1}$ & $\textbf{97.9}$ & $37.4$ & $\textbf{92.9}$ & $\textbf{50.3}$ & $\textbf{55.3}$ & $\textbf{64.9}$ & $50.4$ & $75.0$ & $61.1$ & $\textbf{47.5}$ & $\textbf{47.6}$ & $\textbf{64.3}$ \\
        \bottomrule
    \end{tabular}}
    
    \vspace{5pt}
    \caption{\textbf{Per-class results for Recognition Quality on Cityscapes validation dataset (short-term).} }
    \label{tab:per_class_rq}
\end{table*}
\begin{table*}[t]
    \centering
    \newcommand*\rot{\rotatebox{90}}
    
    \resizebox{1.0\textwidth}{!}{
    \begin{tabular}{l|ccccccccccccccccccc|c}
    \toprule
        & \rot{road} & \rot{sidewalk} & \rot{building} & \rot{wall} & \rot{fence} & \rot{pole} & \rot{traffic light} & \rot{traffic sign} & \rot{vegetation} & \rot{terrain} & \rot{sky} & \rot{person} & \rot{rider} & \rot{car} & \rot{truck} & \rot{bus} & \rot{train} & \rot{motorcycle} & \rot{bicycle} & \rot{mean} \\
        \midrule
        \textbf{Deeplab (Oracle)$\dagger$} & $99.9$ & $91.3$ & $97.8$ & $39.5$ & $52.1$ & $86.1$ & $75.6$ & $93.0$ & $98.5$ & $47.7$ & $94.9$ & $66.9$ & $65.5$ & $80.0$ & $58.2$ & $73.4$ & $43.1$ & $59.9$ & $61.2$ & $72.9$ \\
        \midrule
        \textbf{Deeplab (Last seen frame)} & $99.7$ & $47.6$ & $79.3$ & $12.1$ & $17.0$ & $7.0$ & $12.2$ & $10.1$ & $77.4$ & $13.7$ & $74.1$ & $8.3$ & $4.2$ & $19.8$ & $30.6$ & $38.5$ & $13.6$ & $8.3$ & $4.7$ & $30.4$ \\
        Flow & $99.7$ & $52.2$ & $87.1$ & $11.7$ & $23.4$ & $7.2$ & $17.3$ & $16.5$ & $87.4$ & $18.0$ & $82.1$ & $9.2$ & $4.2$ & $27.5$ & $30.6$ & $37.3$ & $17.8$ & $19.6$ & $8.4$ & $34.6$\\
        Hybrid \cite{terwilliger2019recurrent} (bg) and \cite{luc2018predicting} (fg) & $\textbf{99.9}$ & $64.5$ & $90.4$ & $18.7$ & $22.7$ & $2.1$ & $12.7$ & $17.4$ & $90.5$ & $21.4$ & $84.4$ & $12.5$ & $7.7$ & $38.2$ & $46.2$ & $57.9$ & $37.8$ & $13.4$ & $10.1$ & $39.4$\\
        IndRNN-Stack & $99.7$ & $\textbf{71.2}$ & $\textbf{93.7}$ & $\textbf{26.6}$ & $29.8$ & $14.7$ & $29.2$ & $\textbf{43.8}$ & $\textbf{95.1}$ & $\textbf{27.3}$ & $\textbf{87.6}$ & $25.2$ & $19.5$ & $45.4$ & $47.0$ & $69.6$ & $\textbf{40.0}$ & $22.6$ & $20.0$ & $47.8$\\
        \textbf{Ours} & $99.7$ & $71.1$ & $\textbf{93.7}$ & $26.2$ & $\textbf{31.4}$ & $\textbf{15.5}$ & $\textbf{29.3}$ & $43.5$ & $\textbf{95.1}$ & $27.1$ & $\textbf{87.6}$ & $\textbf{30.8}$ & $\textbf{23.9}$ & $\textbf{52.9}$ & $\textbf{50.7}$ & $\textbf{71.7}$ & $38.7$ & $\textbf{29.2}$ & $\textbf{22.4}$ & $\textbf{49.5}$ \\
        \bottomrule
    \end{tabular}}
    
    \vspace{5pt}
    \caption{\textbf{Per-class results for Recognition Quality on Cityscapes validation dataset (mid-term)}. }
    \label{tab:per_class_rq_mid}
\end{table*}

\end{document}